\documentclass{article}

\usepackage[letterpaper,margin=3cm,footskip=30pt]{geometry}
\usepackage{amsmath}
\usepackage{amssymb}
\usepackage{amsfonts}
\usepackage{amsthm}
\usepackage{paralist}
\usepackage{booktabs}
\usepackage{graphicx}
\usepackage{array}
\usepackage{multirow}
\usepackage{multicol}
\usepackage{comment}
\usepackage{xcolor}
\usepackage{xspace}
\usepackage{hyperref}
\usepackage{balance}
\usepackage{caption}
\usepackage{float}
\usepackage{subcaption}
\usepackage{url}
\usepackage[linesnumbered,ruled,vlined]{algorithm2e}
\usepackage{listings}
\usepackage[most]{tcolorbox}
\usepackage[framemethod=TikZ]{mdframed}
\usepackage{authblk}
\usepackage{soul}
\usepackage{bbding}
\usepackage{lipsum}
\usepackage{url}

\newcommand{\str}[1]{\texttt{#1}}

\newcommand{\mycomment}[1]{}

\newcommand{\myparagraph}[1]{\vspace{1ex}\noindent\underline{\it #1.}\xspace}

\definecolor{lightred}{RGB}{255, 200, 200}
\definecolor{lightgreen}{RGB}{200, 255, 200}
\definecolor{deepviolet}{RGB}{96, 0, 96}

\tcbset{
  promptbox/.style={
    top=10pt,
    colback=white,
    colframe=violet,
    colbacktitle=deepviolet,
    fontupper=\itshape,
    enhanced,
    center,
    attach boxed title to top left={yshift=-0.1in,xshift=0.15in},
    boxed title style={boxrule=0pt,colframe=white},
  }
}
\newtcolorbox{prompt}[2][]{promptbox,title=#2,#1}

\mdfsetup{%
  font=\small,
  innertopmargin=2pt,
  innerbottommargin=2pt,
  innerleftmargin=8pt,
  innerrightmargin=8pt,
  frametitleaboveskip=2pt,
  frametitlebelowskip=1pt,
  frametitlealignment=\center,
  middlelinecolor=blue
}
\newenvironment{snippet}
{
  \vspace{1ex}
  \begin{mdframed}
  \itshape
}
{
  \end{mdframed}
  \vspace{1ex}
}

\newcommand{\jellyfish}{Jellyfish\xspace}

\newcommand{\llama}{Llama\xspace}

\newcommand{\llamatwo}{Llama 2\xspace}

\newcommand{\holoclean}{HoloClean\xspace}

\newcommand{\magellan}{Magellan\xspace}

\SetFuncSty{text}
\SetArgSty{text}
\SetKw{OR}{or}
\SetKw{AND}{and}
\SetKw{NOT}{not}
\SetKw{TRUE}{true}
\SetKw{FALSE}{false}
\SetKw{NULL}{nil}
\SetKw{Break}{break}
\SetKw{Continue}{continue}
\SetKwInOut{Input}{Input}
\SetKwInOut{Output}{Output}

\title{Jellyfish: A Large Language Model for Data Preprocessing}

\date{}

\begin{document}

\newcommand{\osaka}{$^{1}$}
\newcommand{\nec}{$^{2}$}
\newcommand{\nagoya}{$^{3}$}
\newcommand{\osanag}{$^{1,3}$}

\author{\osaka Haochen Zhang, \nec Yuyang Dong, \osanag Chuan Xiao, \nec Masafumi Oyamada\\
\small{\osaka Osaka University, \nec NEC Corporation, \nagoya Nagoya University}\\
\small{\{chou.koushin, chuanx\}@ist.osaka-u.ac.jp, \{dongyuyang, oyamada\}@nec.com}
}

\def\thefootnote{}\footnotetext{Haochen Zhang and Yuyang Dong are co-first authors who contributed equally to this work. Chuan Xiao is the corresponding author.}

\def\thefootnote{}\footnotetext{Our models are available at: \url{https://huggingface.co/NECOUDBFM/Jellyfish} . Our instruction dataset is available at: \url{https://huggingface.co/datasets/NECOUDBFM/Jellyfish-Instruct} .}

\maketitle

\begin{abstract}
  This paper explores the utilization of LLMs for data preprocessing (DP), a crucial step in the data mining pipeline that transforms raw data into a clean format conducive to easy processing. Whereas the use of LLMs has sparked interest in devising universal solutions to DP, recent initiatives in this domain typically rely on GPT APIs, raising inevitable data breach concerns. Unlike these approaches, we consider instruction-tuning local LLMs (7 -- 13B models) as universal DP task solvers that operate on a local, single, and low-priced GPU, ensuring data security and enabling further customization. We select a collection of datasets across four representative DP tasks and construct instruction tuning data using data configuration, knowledge injection, and reasoning data distillation techniques tailored to DP. By tuning Mistral-7B, Llama 3-8B, and OpenOrca-Platypus2-13B, our models, namely, \jellyfish-7B/8B/13B, deliver competitiveness compared to GPT-3.5/4 models and strong generalizability to unseen tasks while barely compromising the base models' abilities in NLP tasks. Meanwhile, \jellyfish offers enhanced reasoning capabilities compared to GPT-3.5. 
\end{abstract}
\section{Introduction}
\label{sec:intro}
The proliferation of large language models (LLMs) has catalyzed a diverse array of applications, extending beyond the domain of NLP to encompass a wide range of fields that require the processing of natural language data. Notably, LLMs have been applied in areas such as software engineering~\cite{qian2023communicative,tang2024collaborative}, computer simulation~\cite{wu2023smart,gao2023large}, data analytics~\cite{cheng2023gpt,savelka2023can}, and tabular data processing~\cite{li2023table,lu2024large,zhang2023tablellama}. 

This paper focuses on the utilization of LLMs for data preprocessing (DP), a critical step in the data mining pipeline that involves transforming raw data into a manageable and processable format ready for use. Over the past decades, significant strides have been made in various DP tasks. Until 2021, most efforts were concentrated on one or two specific tasks such as error detection (ED)~\cite{heidari2019holodetect,mahdavi2019raha}, data imputation (DI)~\cite{rekatsinas2017holoclean,mahdavi2020baran,mei2021capturing}, schema matching (SM)~\cite{zhang2021smat}, and entity matching (EM)~\cite{konda2016magellan,li2020deep}. A key challenge in developing generic solutions to DP is that these tasks differ in nature: they deal with errors, anomalies, matches, etc. and require different actions such as detection, repairing, and alignment. 


With the advent of LLMs like GPT-3 and subsequent versions, researchers have found a key to address this challenge, spurring the development of generic solutions for a wider array of DP tasks~\cite{narayan2022can,zhang2023large}. The application of LLMs in DP has the following strengths: 
\begin{inparaenum} [(1)]
  \item The primary strengths of using LLMs in DP lie in their ability to process natural language. Most LLMs provide a  prompting interface with which users can interact and assign tasks in natural language, contrasting with existing DP solutions that require computer programming or specific tools (e.g., \holoclean~\cite{rekatsinas2017holoclean} and \magellan~\cite{konda2016magellan}).
  \item With the knowledge acquired through training on vast amounts of data, LLMs are universal problem solvers capable of identifying errors, anomalies, and matches in data (and particularly unseen datasets in unseen tasks), aligning with the aims of DP tasks without needing human-engineered rules~\cite{razniewski2021language}. 
  \item LLMs are excellent reasoners~\cite{kojima2022large}, enabling them to not only return DP results but also provide the reasons for these results. In this sense, their answers are more interpretable than those of other deep learning approaches.
  \item LLMs can be conditioned by few-~\cite{brown2020language} or zero-shot~\cite{kojima2022large} prompting. As such, we can condition the criteria for DP tasks (e.g., the degree of matching) using few-shot examples or zero-shot prompts, contrasting with traditional solutions based on a threshold~\cite{sagi2013schema,konda2016magellan} or a time-consuming training process to fit to the data~\cite{mei2021capturing}.
\end{inparaenum}

Despite these strengths, existing LLM-based solutions to DP~\cite{narayan2022can,zhang2023large,korini2023column}, with reliance on GPT APIs, have raised concerns about data breaches, as evidenced by OpenAI's first confirmed data breach involving ChatGPT~\cite{gpt-data-breach}. Another limitation is the difficulty in domain specification~\cite{narayan2022can}. When dealing with data from highly specialized domains, training the LLMs used in these solutions can be costly (e.g., GPT-3.5) and even unavailable due to frozen parameters (e.g., GPT-4), posing difficulty in customizing the model. 



\begin{figure*}[!t]
  \centering
  \includegraphics[width=\linewidth]{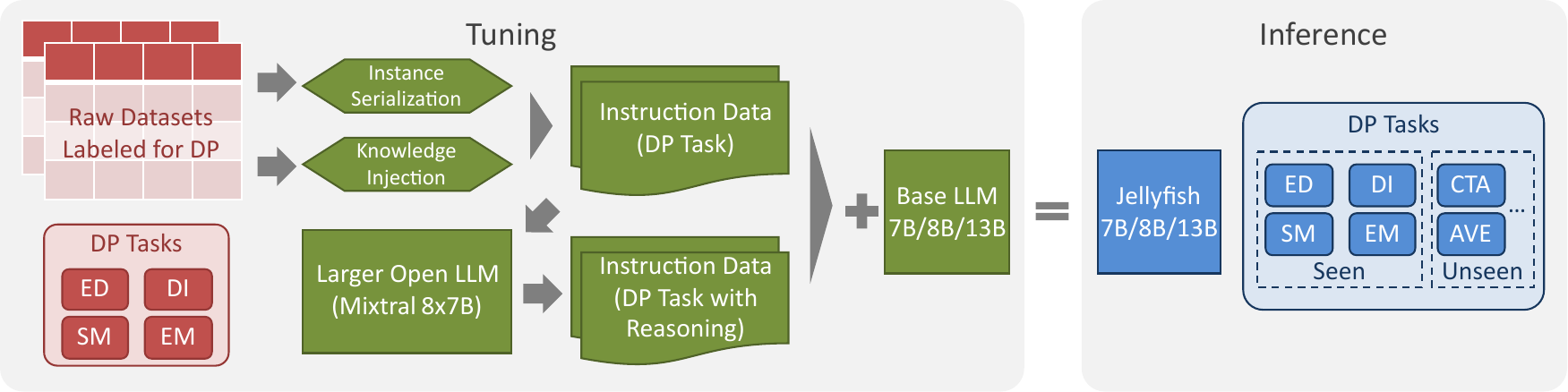}
  \caption{Overview of instruction tuning for data preprocessing.}
  \label{fig:overview}
\end{figure*}

In response to these challenges, we propose to construct instruction data and tune LLMs for various DP tasks. The tuned model, namely \jellyfish, distinguish itself with several key features:
\begin{itemize} 
    \item \textbf{Versatility:} \jellyfish is a universal DP task solver tuned to the following tasks: ED and DI for data cleaning, and SM and EM for data integration. 
    \item \textbf{Cost-Efficiency and Security:} Varying from 7B to 13B, \jellyfish can operate on a local, single, and low-priced GPU, ensuring data security and allowing further tuning.
    \item \textbf{Customizability:} Capable of understanding natural language, \jellyfish allows users to manually craft instructions for DP tasks (or simply use our prompts in this paper) and apply prompt engineering techniques to tailor it to specific tasks and datasets.
    \item \textbf{Domain Knowledge:} Unlike many existing methods that rely heavily on handcrafted knowledge during inference~\cite{rekatsinas2017holoclean,qin2023bclean}, \jellyfish features domain knowledge in its instruction tuning and enables optional knowledge injection during inference.
    \item \textbf{Interpretability:} By employing reasoning data in its instruction tuning, \jellyfish's interpretation ability provides natural language explanations of its outputs.
\end{itemize}


Whereas instruction tuning of LLMs has been largely used for unstructured text~\cite{zhang2023instruction}, the construction of \jellyfish is non-trivial in the sense that (1) it tunes for structured data, (2) it finds a good data configuration for various DP tasks, and (3) it specifies domain knowledge that can be applied to unseen datasets. Besides, it is expected that the model's performance in NLP tasks can be preserved for generalizability and further customization. To the best of our knowledge, this is the first study that investigates instruction tuning for DP with LLMs as universal solutions. 

As depicted in Figure~\ref{fig:overview}, \jellyfish is constructed by carefully selecting data from several public datasets widely used for DP evaluation, considering their impacts on the overall performance. By instance serialization, raw data is serialized into instruction tuning prompts. By knowledge injection, task- and dataset-specific knowledge -- particularly domain knowledge that can be extended to unseen datasets -- is infused to the prompts. Moreover, we resort to Mixtral-8x7B-Instruct-v0.1 to generate reasoning data. As such, \jellyfish distills Mixtral's knowledge in reasoning DP results. 

Our evaluation focuses on tuning a set of prevalent open LLMs, including Mistral-7B-Instruct-v0.2 (as \jellyfish-7B), Llama 3-8B (as \jellyfish-8B), and OpenOrca-Platypus2-13B (as \jellyfish-13B). The results show that our instruction data applies to all these base models, substantially improving the DP performance. Compared to two categories of baseline methods, (1) non-LLM methods -- typically solutions based on machine learning (ML) or pre-trained language models (PLMs) -- and (2) LLM methods -- typically GPT series methods, \jellyfish-13B consistently outperforms non-LLM methods on its seen datasets, and its effectiveness on unseen datasets even surpasses non-LLM methods on their respective seen datasets. Meanwhile, \jellyfish-7B/8B also exhibit competitiveness, especially on DI and EM tasks. For unseen tasks, \jellyfish models also deliver strong performance, rivaling GPT-3.5/4 models and showcasing generalizability to a wider range of DP tasks beyond the four tasks used for tuning. Our evaluation reveals the impacts of data configuration and the use of reasoning data in building \jellyfish, and discovers that \jellyfish barely compromises the base model's NLP performance. Furthermore, experiments demonstrate the advantage of \jellyfish's interpretation over GPT-3.5 in reasoning capabilities as well as the effectiveness of knowledge injection.

Our contributions are summarized as follows.
\begin{itemize}
    \item We develop \jellyfish, an instruction-tuned LLM as a universal DP task solver. 
    \item \jellyfish showcases several notable features: universal model design, moderate model size, assurance of data security, feasibility for further tuning, natural language instruction handling, optional specification of prior knowledge, and model interpretability.
    \item Our experiments demonstrate \jellyfish-7B, 8B, and 13B models' effectiveness in DP task solving, generalizability to new tasks beyond what they are tuned for, and superior reasoning abilities.
\end{itemize}

The rest of the paper is organized as follows: Section~\ref{sec:prelim} introduces the DP tasks targeted by our model and briefly reviews LLMs. Section~\ref{sec:tuning} describes the instruction data for tuning \jellyfish. Section~\ref{sec:dp} introduces how to use \jellyfish for solving DP tasks. Section~\ref{sec:extend} discusses the extensions to unseen tasks. Section~\ref{sec:exp} reports experimental results and analysis. Section~\ref{sec:related} reviews related works on DP. Section~\ref{sec:concl} concludes this paper. 
\section{Preliminaries}
\label{sec:prelim}

\subsection{Data Preprocessing}
\label{sec:prelim:problem}
In data mining, DP is a crucial step that deals with noise, missing values, inconsistencies, and heterogeneity in data. Major DP procedures include data cleaning, data integration, data transformation, and data reduction~\cite{han2022data}. In this initial exploration of LLMs for DP, we concentrate on tabular data, one of the most common data types. 

Our data model operates on relational tables specified by schemas. We assume all attributes are either numerical (including binary) or textual (including categorical) values. Diverging from the traditional definition that presents the entire dataset and finds or fixes all the errors (or matches, etc.) within, we define the problem by handling one record (or a pair, depending on the task) at a time, so the prompt can be easily written and its length is within LLMs' token limitation. Next, we outline the DP tasks involved in this study: 

\begin{itemize} 
    \item \textbf{Error Detection (ED)}: Given a record (i.e., a tuple in a relational table) and an attribute, our task is to detect whether there is an error in the cell value of this attribute.
    \item \textbf{Data Imputation (DI)}: Given a record and an attribute such that cell value for this attribute is missing, our task is to infer its correct value. 
    \item \textbf{Schema Matching (SM)}: Given a pair of attributes represented in the form of (name, description), our task is to find whether they refer to the same attribute.
    \item \textbf{Entity Matching (EM)}: Given a pair of records, our task is to infer whether they refer to the same entity. 
\end{itemize}


These four tasks form the most critical part of DP~\cite{narayan2022can,zhang2023large} and are extensively discussed in the context of data mining~\cite{han2022data}. We use them for instruction tuning. Besides, we consider two unseen tasks: 
\begin{itemize} 
    \item \textbf{Column Type Annotation (CTA)}: Given a table with no header, our task is to infer the type of each column from a set of predefined types (e.g., name, time, location).
    \item \textbf{Attribute Value Extraction (AVE)}: Given a text description of an entity and a set of predefined attributes, the task is to extract attribute values from the text description. 
\end{itemize}

We term each input object an \emph{instance}, i.e., a record for ED and DI, a pair of attributes for SM, a pair of records for EM, a table or a column for CTA, and a text description for AVE.  

\subsection{Large Language Models}
\label{sec:prelim:llm}
With advancements in the field of natural language processing (NLP), LLMs have become one of the hottest topics in the AI research community. Representative LLMs include OpenAI's GPT series (in particular, GPT-3, 3.5, and 4), Anthropic's Claude, Google's Gemini, Mistral AI's Mistral~\cite{jiang2023mistral}, Meta's \llama~\cite{touvron2023llama} series, as well as their variants that can be found at Hugging Face~\cite{llamavariants}. Due to their superb ability to process natural language, LLMs have not only been used in NLP applications (e.g., ChatGPT and Claude), but also catalyzed the rise of LLM-powered autonomous agents~\cite{wang2023survey} as AI assistants (e.g., by GPTs) or tools for engineering~\cite{qian2023communicative,hong2023metagpt} or simulation~\cite{xi2023rise,wu2023smart} purposes. Another popular LLM-centric research direction is retrieval-augmented generation (RAG) \cite{lewis2020retrieval,li2022survey}, which gives LLMs access to external information to improve generation performance. We refer readers to \cite{zhao2023survey} for a survey on LLMs. Some LLMs are open-source (e.g., \llama and \llamatwo), and they can be fine-tuned with additional tasks to improve their abilities in logical reasoning, question answering, and so on. Among these fine-tuning approaches, instruction tuning~\cite{zhang2023instruction} has become a prevalent one which further trains LLMs on a dataset consisting of (instruction, output) pairs in a supervised fashion, hence bridging the gap between the next-word prediction objective of LLMs and the users' objective of having LLMs adhere to human instructions. For efficiency of fine-tuning, parameter-efficient fine-tuning (PEFT) approaches enable adaptation of LLMs to downstream applications without fine-tuning all the parameters. Notable methods are adapter tuning~\cite{houlsby2019parameter}, prefix-tuning~\cite{li2021prefix}, and low-rank adaptation (LoRA)~\cite{hu2021lora}. In particular, LoRA achieves significantly fewer trainable parameters and no additional inference latency, and has become a prevalent PEFT approach. 

In addition to the strengths outlined in Section~\ref{sec:intro}, we discuss the limitations of LLMs in the context of DP: 
\begin{inparaenum} [(1)]
  \item LLMs often require substantial computational resources, thereby increasing the cost of use and compromising the efficiency and scalability of DP on large-scale data.
  \item Due to token limitation (the maximum input length, e.g., 4k tokens for GPT-3.5) and lack of memory for keeping historical information, the input to the LLM is often instance-by-instance, and the DP results may exhibit inconsistency across different instances. Simply raising the token limitation (e.g., 128k tokens for GPT-4-turbo) does not solve the problem, because performance may degrade due to increased lengths of input~\cite{liu2023lost}.
  \item LLMs sometimes exhibit hallucination~\cite{zhang2023siren}, i.e., they generate text that is plausible-sounding but factually incorrect or non-sensical, as they lack a fundamental understanding of the world and rely solely on the patterns they learned during training.
\end{inparaenum}
\section{Instruction Tuning of Jellyfish}
\label{sec:tuning}

\subsection{Dataset Preparation}
\label{sec:tuning:prepare}
For the four seen tasks, we choose a series of datasets that have been widely used in previous studies and cover a variety of application domains.
\begin{inparaenum} [(1)]
    \item ED: Adult and Hospital~\cite{heidari2019holodetect}; 
    \item DI: Buy and Restaurant~\cite{mei2021capturing}; 
    \item SM: MIMIC-III and Synthea~\cite{zhang2021smat};  
    \item EM: Amamzon-Google, Beer, DBLP-ACM, DBLP-GoogleScholar, Fodors-Zagats, and iTunes-Amazon from the Magellan data repository~\cite{magellandata}. 
\end{inparaenum}
We use the publicly available version of these datasets~\cite{narayan2022can}, where errors and missing values are already injected to the datasets of ED and DI, respectively. 

To determine the data size for each task, we first consider a constraint that for fair comparison with non-LLM methods~\cite{mei2021capturing,zhang2021smat,li2020deep}, the training data in building \jellyfish does not exceed those used for building these methods, which serve as a pool of 115k instances. Then, we have the following observations (Section~\ref{sec:exp:impact-configuration}): 
\begin{inparaenum} [(1)]
    \item The performance of DI can benefit from the other three tasks, but increasing DI data is relatively negative to them. 
    \item Increasing ED and SM data is generally beneficial to other tasks.
    \item Increasing SM data is beneficial to the overall DP performance.
    \item Increasing EM data compromises the performance of other tasks, but keeping its size is the key to the EM performance. 
\end{inparaenum}

Based on these observations, we use all the ED and DI data in the 115k pool as their sizes are small, and then choose a large data size for SM and a moderate data size for EM. Specifically, we control the data used in large EM datasets (e.g., for DBLP-GoogleScholar, 1/3 is chosen from the pool). As such, we determine the data size for the four tasks, as shown in Table~\ref{tab:datasets-dptuning}. 

In addition, we undertake the following efforts to prepare data:
\begin{inparaenum} [(1)]
    \item Given the disproportionately small number of positive instances compared to negative ones, we incorporate all positive instances available in the datasets.
    \item For ED, since missing values can be interpreted as either errors or non-errors, depending on the context, we create two versions of each instance: one treating missing values as errors and the other as non-errors. The duplication is guided by knowledge injection, which is to be detailed in Section~\ref{sec:tuning:solver}. 
\end{inparaenum}

Next, we transform raw data to (1) DP task data, for DP task-solving ability, and (2) DP task with reasoning data, for interpretation ability. They can be jointly used for tuning a \jellyfish model. 

\begin{table}[!t]
    \small
    \caption{DP task data statistics. \#Positives denotes the number of instances having an error (for ED) or matching objects (for SM and EM). $\times$2 denotes duplication of instances for treating missing values as errors or not.}
    \centering
    \begin{tabular}{|c|c|r|r|}
         \hline
         \textbf{Task} & \textbf{Dataset} & \textbf{\#Instances} & \textbf{\#Positives} \\ \hline
         \multirow{2}{*}{ED} & Adult & 550$\times$2 & 35$\times$2 \\
         & Hospital & 1710$\times$2 & 44$\times$2 \\ \hline
         \multirow{2}{*}{DI} & Buy & 586 & N/A \\
         & Restaurant & 778 & N/A \\ \hline
         \multirow{2}{*}{SM} & MIMIC-III & 7000 & 11 \\
         & Synthea & 5000 & 18 \\ \hline
         \multirow{6}{*}{EM} & Amazon-Google & 6874 & 699 \\
         & Beer & 359 & 54 \\
         & DBLP-ACM & 5000 & 885 \\
         & DBLP-GoogleScholar & 5000 & 924 \\
         & Fodors-Zagats & 757 & 88 \\
         & iTunes-Amazon & 430 & 105\\ 
         \hline
    \end{tabular}
    \label{tab:datasets-dptuning}
\end{table}

\begin{figure*}[!t]
  \centering
  \includegraphics[width=\linewidth]{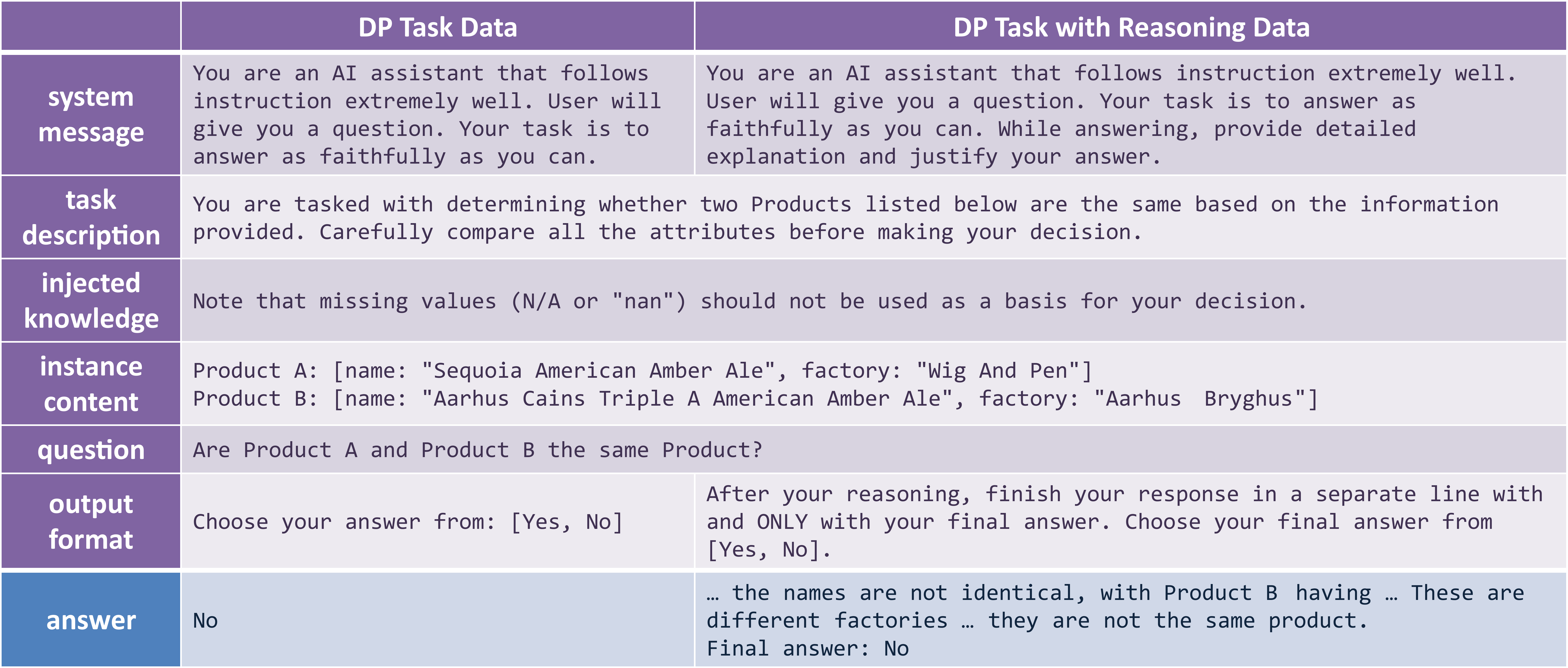}
  \caption{Example prompt in instruction data. The leftmost column is description and not prompted to the model. Response indicates the answer to the prompt. Detailed prompts are provided in Appendix~\ref{sec:app:prompts}.}
  \label{fig:instruction}
\end{figure*}

\subsection{DP Task Data}
\label{sec:tuning:solver}
To prepare the DP task data for an LLM, we need to serialize (a.k.a. contextualize) each instance in the raw data to a prompt. The prompt contains the task description, the instance content, and any injected knowledge. To describe our techniques for constructing the DP task data for training, we use an example for an instance in the Beer dataset used for EM, as shown in Figure~\ref{fig:instruction}. 

At the beginning, there is a system message guiding the model behavior. Here, we instruct the model to act as an AI assistant to answer the user's question, and its response should always respect this constraint. Then, we describe the DP task. The following part refers to injected knowledge. There are two types of injected knowledge: (1) general knowledge that applies to many datasets, and (2) specific knowledge that only applies to the given dataset. In this example, the knowledge belongs to general knowledge and concerns with missing values. Such knowledge injection may prevent the model incorrectly handling certain values in the dataset, especially when training data is noisy. The following part pertain to the instance content. Finally, there is a question presented to the model, and the output format is specified afterwards. 

Whereas in the above example we specify knowledge on missing values, there are other forms of general knowledge used in tuning, including error types and terminology. For example, for ED, we inform the model of the fact that errors can include, but are not limited to, spelling errors, inconsistencies, or values that do not make sense for that attribute; for EM, we instruct the model to consider the full name of an attribute and its acronym to determine if the two values are the same. Specific knowledge highly depends on the application domain, mainly including constraints or rules that pertain to the dataset. For example, in publication datasets, authors' names may occur in different forms and different orders even for the same article. Additionally, the model can be configured to assign greater importance to certain attributes. In the context of product data, for example, the model is directed to prioritize the comparison of product numbers. Specific knowledge can be applicable to datasets within the same domain, thereby enhancing the model's performance on unseen datasets, particularly in scenarios where prior knowledge about these datasets is absent. Overall, the knowledge injected through tuning becomes the built-in knowledge of the model and can be used even without user-specification during inference. 

\subsection{DP Task with Reasoning Data}
\label{sec:tuning:interpreter}

(DP task with) reasoning data, not only empowers the model to interpret the DP results, but also has the potential in enhancing the DP performance in the sense that the model can learn the rationale behind DP, thereby generalizing to unseen scenarios whose underlying logic resembles the tuned tasks/datasets. On the other hand, due to the small size of local LLMs, tuning the model with excessive reasoning data may compromise its ability to conduct the tuned DP tasks. Thus, we need to strike a balance between DP performance and generalizability. In general, we observe that native models (Mistral and Llama 3) are more likely to benefit from the use of reasoning data (Section~\ref{sec:exp:impact-reasoning}). 

Another key feature in our reasoning data is that we resort to a larger open LLM, Mixtral-8x7B-Instruct-v0.1, to retrieve reasoning answers as ground truths. As such, \jellyfish distills Mixtral's knowledge in reasoning for DP. Since this does not involve external APIs like GPT-4, data security can be ensured, in case users want to include confidential information in the reasoning data. 

We use the same set of datasets as DP task data to construct the reasoning data. The prompt in reasoning data only differs from DP task data in the reasoning instructions (Figure~\ref{fig:instruction}, system message and output format). To retrieve reasoning answers from Mixtral, we add a hint at the end of the prompt for the correct DP result (e.g., ``yes/no'' for matching tasks), hence to instruct Mixtral to reason in the right direction (Appendix~\ref{sec:app:prompts:reasoning}). Note that such hint does not appear in the prompt given to \jellyfish. 

To control the size and quality of reasoning data, we select data as follows: 
\begin{inparaenum} [(1)]
    \item For ED and SM, we keep all positive instances due to their small numbers, and then sample negative instances. 
    \item For DI, we keep all instances due to the small data size. 
    \item For EM, we sample instances. 
\end{inparaenum}
From the 115k pool, we tune the numbers in the sample to make four sets of reasoning data with 8k, 11k, 14k, and 20k instances, respectively (Table~\ref{tab:datasets-reasoning}). Moreover, from the answers returned by Mixtral, we remove low-quality ones that simply rephrase instance contents, as they barely refers to reasoning. 

\begin{table}[!t]
    \small
    \caption{Statistics of reasoning data for instruction tuning. We report the numbers of instances for each task. For ED and DI, the numbers refer to the amount after duplicating the instances having missing values.}
    \centering
    \begin{tabular}{|c|c|c|c|c|c|}
         \hline
         \multirow{2}{*}{\textbf{Dataset}} & \multicolumn{5}{c|}{\textbf{Task}} \\ \cline{2-6}
         & ED & DI & SM & EM & Total \\ \hline
         reasoning-8k  & 3056 & 1364 & 2000 & 2000 & 8420 \\ \hline
         reasoning-11k & 3056 & 1364 & 3500 & 3500 & 11420 \\ \hline
         reasoning-14k & 3056 & 1364 & 5000 & 5000 & 14420 \\ \hline
         reasoning-20k & 3056 & 1364 & 8600 & 7000 & 20020 \\
         \hline
    \end{tabular}
    \label{tab:datasets-reasoning}
\end{table}

\section{Inference with Jellyfish}
\label{sec:dp}
For inference, the prompt is same as the instruction data shown in Figure~\ref{fig:instruction}. Users can craft dataset-specific knowledge into the prompt, such as the domain knowledge (e.g., constraints) outlined in the previous section. Such user-specified knowledge is optional. 

\myparagraph{Feature Engineering}
Users can optionally select a subset of features to improve performance. For instance, for EM in the Beer dataset, \str{name} and \str{factory} are more relevant features, while \str{style} and \str{ABV} are less relevant. Hence users may choose to use only \str{name} and \str{factory} as attributes. Such feature engineering can be also implemented in the prompt as specific knowledge, e.g., \textit{you should only consider name and factory and ignore other attributes}.

\myparagraph{Prompt Engineering}
Prompt engineering~\cite{prompt-engineering} is the process of structuring text to enhance the model performance. We incorporate few-shot prompting~\cite{brown2020language}, which conditions the \jellyfish models to learn from a small selection of examples drawn from the dataset. The prompts for few-shot examples are reported in Appendix~\ref{sec:app:few-shot}.



\myparagraph{Batch Processing}
To enable Jellyfish models to perform inference in batches rather than processing single instances individually, we can employ prefix caching~\cite{kwon2023efficient}, available in the vLLM~\cite{vllm} library, because the instructions for the batch share the same prefix, only differing in the instance content.

\section{Extensions to Unseen Tasks}
\label{sec:extend}
For unseen tasks, we consider two case studies: CTA and AVE, as outlined in Section~\ref{sec:prelim:problem}. \jellyfish models can be easily extended to support them by employing the prompt engineering techniques in existing LLM-based solutions, hence simplifying its use in unseen tasks. 

\myparagraph{Column Type Annotation}
As a task in the realm of table understanding, CTA is an essentially DP step for data search~\cite{chapman2020dataset}, knowledge base completion~\cite{ritze2016profiling}, and data integration a data lake~\cite{hai2023data}. We follow the two-stage pipeline proposed in \cite{korini2023column}, which was designed for ChatGPT and based on chain-of-thought~\cite{wei2022chain}, a technique that enables complex reasoning capabilities through intermediate reasoning steps. 

Given a table to be annotated, in the first stage, the model predicts the domain of the table. In the second stage, given a set of predefined types, the model determines the type of column based on sample values extracted from it. The chain-of-thought prompt instructs the model in a step-by-step manner. For example, to predict the domain of the table, there are four steps: 
\begin{inparaenum} [(1)]
    \item look at the input and make a table out of it, 
    \item look at the cell values in detail, 
    \item decide if the table describes domain A, domain B ... and 
    \item answer with the domain. 
\end{inparaenum}
Then, the model follows this prompt to cope with the task. The column type selection in the second stage works in the same way, except that table is replaced by column and domains are replaced by candidate types. 

\myparagraph{Attribute Value Extraction}
Given a text description, AVE is an information extraction task that discovers missing values of attributes and reconstructs a table. For this task, we follow the prompt in \cite{brinkmann2023product} designed for GPT-4. The prompt is simple, beginning with the task description. Then, the instance content follows, with the description of the entity and the attribute to be extracted. Finally, an exception rule is mentioned: if the attribute cannot be extracted, the model should answer ``N/A''.

We also would like to mention that \jellyfish models enable further fine-tuning. Users may choose to condition the model for specific DP tasks or domains to seek better performance. Moreover, \jellyfish models can be utilized for multiple tasks in a DP pipeline, e.g., data cleaning followed by data integration on the same sets of data. It is likely that the DP tasks within this pipeline belong to the same domain. In this case, \jellyfish models may deliver consistency in handling the data in different tasks due to the built-in domain knowledge acquired through instruction tuning for DP. 

\section{Experiments}
\label{sec:exp}

\subsection{Experimental Setup}
\label{sec:exp:setup}

\myparagraph{Datasets}
Apart from the seen datasets in \jellyfish (Section~\ref{sec:tuning}), we use the following datasets as unseen data, where CTA and AVE are used for case studies on unseen tasks. 
\begin{inparaenum} [(1)]
    \item ED: Flights and Rayyan~\cite{mahdavi2019raha}; 
    \item DI: Flipkart~\cite{flipkart} and Phone~\cite{phone} from Kaggle; 
    \item SM: CMS~\cite{zhang2021smat}; 
    \item EM: Abt-Buy and Walmart-Amazon from the Magellan data repository~\cite{magellandata}; 
    \item CTA: SOTAB~\cite{korini2023column}; 
    \item AVE: AE-110k and OA-Mine~\cite{brinkmann2023product}. 
\end{inparaenum}
The statistics of the datasets are reported in Table~\ref{tab:datasets-test}. We generate train/valid/test splits following the protocols for Adult and Hospital~\cite{heidari2019holodetect},  Flipkart and Phone~\cite{mei2021capturing}, and MIMIC-III and CMS~\cite{zhang2021smat}. The other datasets have already been provided with splits~\cite{narayan2022can,korini2023column,brinkmann2023product}. A subset of the train/valid splits is used in \jellyfish, as reported in Tables~\ref{tab:datasets-dptuning} and~\ref{tab:datasets-reasoning}. 


\begin{table}[!t]
    \small
    \centering
    \caption{Testing dataset statistics. For Walmart-Amazon, the entities belong to a different category of products from the Amazon dataset used for instruction tuning.}
    \begin{tabular}{|c|c|c|r|}
         \hline
         \textbf{Task} & \textbf{Type} & \textbf{Dataset} & \textbf{\#Instances} \\ \hline
         \multirow{4}{*}{ED} & \multirow{2}{*}{Seen} & Adult & 9900 \\
         & & Hospital & 17101 \\ \cline{2-4}
         & \multirow{2}{*}{Unseen} & Flights & 12832 \\
         & & Rayyan & 8997 \\ \hline
         \multirow{4}{*}{DI} & \multirow{2}{*}{Seen} & Buy & 65 \\
         & & Restaurant & 86 \\ \cline{2-4}
         & \multirow{2}{*}{Unseen} & Flipkart & 2675 \\
         & & Phone & 1194 \\ \hline
         \multirow{3}{*}{SM} & \multirow{2}{*}{Seen} & MIMIC-III & 6408 \\
         & & Synthea & 2964 \\ \cline{2-4}
         & Unseen & CMS & 2564 \\ \hline
         \multirow{8}{*}{EM} & \multirow{6}{*}{Seen} & Amazon-Google & 2293 \\
         & & Beer & 91 \\
         & & DBLP-ACM & 2473 \\
         & & DBLP-GoogleScholar & 5742 \\
         & & Fodors-Zagats & 189 \\
         & & iTunes-Amazon & 109 \\ \cline{2-4}
         & \multirow{2}{*}{Unseen} & Abt-Buy & 1946 \\
         & & Walmart-Amazon & 2049 \\ \hline
         CTA & Unseen & SOTAB & 250 \\ \hline
         \multirow{2}{*}{AVE} & \multirow{2}{*}{Unseen} & AE-110K & 1482 \\
         & & OA-Mine & 2451 \\ 
         \hline
    \end{tabular}
    \label{tab:datasets-test}
\end{table}

\myparagraph{Jellyfish Models}
We instruction-tune three base models: 
\begin{inparaenum} [(1)]
    \item Mistral-7B (Mistral-7B-Instruct-v0.2~\cite{jiang2023mistral}), 
    \item Llama 3-8B (Llama-3-8B-Instruct~\cite{llama3}), and 
    \item OOP2-13B (OpenOrca-Platypus2-13B~\cite{hunterlee2023orcaplaty1}), a \llamatwo-13B variant with enhanced reasoning capabilities and logic proficiency. 
\end{inparaenum}
The tuned models are dubbed \jellyfish-7B, \jellyfish-8B, and \jellyfish-13B, respectively. The 7B and 8B models are tuned with both DP task and reasoning data (15k reasoning instances for the 7B model and 8k for the 8B model). The 13B model is tuned with only DP task data. As such, \jellyfish-7B and \jellyfish-8B are interpretation models while \jellyfish-13B is a task solver dedicated to the tuned tasks. 

We report hyperparameter setup in Appendix~\ref{sec:app:exp-setup-detail} and injected knowledge in Appendix~\ref{sec:app:injected-knowledge}. For inference, the (zero-shot) prompts are the same as DP task data and reasoning data, respectively. We apply general knowledge in the prompts, e.g., missing values in matching tasks and error types in ED. Dataset-specific knowledge is not used. When few-shot prompting is enabled, we equip LLMs with three examples for each dataset, covering both positives and negatives (Appendix~\ref{sec:app:few-shot}). 

\myparagraph{Baselines}
We categorize existing methods into non-LLM methods and LLM methods. For non-LLM methods, we select the following baselines, in line with \cite{narayan2022can}: 
\begin{inparaenum} [(1)]
    \item ED: HoloDetect~\cite{heidari2019holodetect} and Raha~\cite{mahdavi2019raha}; 
    \item DI: IPM~\cite{mei2021capturing}; 
    \item SM: SMAT~\cite{zhang2021smat}; 
    \item EM: Ditto~\cite{li2020deep} and Unicorn~\cite{tu2023unicorn}; 
    \item CTA: RoBERTa~\cite{liu2019roberta}. 
\end{inparaenum}
For their performance, we follow the best numbers reported in prior works~\cite{narayan2022can,korini2023column,tu2023unicorn}. Other methods such as Baran~\cite{mahdavi2020baran}, HoloClean~\cite{rekatsinas2017holoclean}, and DODUO~\cite{suhara2022annotating}, have been shown to be outperformed by the above competitors~\cite{mei2021capturing,narayan2022can,korini2023column}, and hence are not compared here. 

LLM methods are GPT-3 (\texttt{text-davinci-002}), GPT-3.5 (\texttt{gpt-3.5-turbo-0301}), Table-GPT~\cite{li2023table} (GPT-3.5 fine-tuned for tables), GPT-4 (\texttt{gpt-4-0314}), GPT-4o (\texttt{gpt-4o-2024-05-13}), Stable Beluga 2 70B~\cite{stable-beluga-2}, and SOLAR 70B~\cite{solar}. We follow the numbers reported in previous works~\cite{narayan2022can,zhang2023large,brinkmann2023product}. Few-shots are used in line with \jellyfish models for fair comparison. TableLlama~\cite{zhang2023tablellama}, which can handle CTA, is not compared because it is tuned for CTA, whereas our purpose is to evaluate the performance of LLMs on CTA as an unseen task.



\myparagraph{Metrics}
For DP task solving, we measure accuracy for DI, F1 score for ED, DI, EM, and AVE, and micro-F1 for CTA, all reported on a 100-scale. 

\myparagraph{Environment}
Training and inference of LLMs are conducted on NVIDIA A100 80GB GPUs. 
We employ LoRA~\cite{hu2021lora} and FlashAttention-2~\cite{dao2023flashattention} for tuning and vLLM with PageAttention~\cite{kwon2023efficient} for inference. 

\begin{table*}[!t]
    \small
    \centering
    \caption{DP performance on seen tasks, accuracy for DI and F1 score for the other three tasks, with winners in boldface and runners-up underlined. All datasets are seen for non-LLM methods and Table-GPT. All datasets are unseen for GPT-3/3.5/4/4o. For LLM methods, zero-shot is used on seen datasets and few-shot is used on unseen datasets. ``--'' indicates numbers not reported in prior works for this dataset.}
    \resizebox{\linewidth}{!}{
    \begin{tabular}{| c | c | c | p{.08\linewidth}<{\centering} | p{.08\linewidth}<{\centering} | p{.08\linewidth}<{\centering} | p{.08\linewidth}<{\centering} | p{.08\linewidth}<{\centering} | p{.08\linewidth}<{\centering} | p{.08\linewidth}<{\centering} | p{.08\linewidth}<{\centering} | p{.08\linewidth}<{\centering} |}
         \hline
         \multirow{2}{*}{\textbf{Task}} & \multirow{2}{*}{\textbf{Type}} & \multirow{2}{*}{\textbf{Dataset}} & \multicolumn{9}{c|}{\textbf{Model}} \\ \cline{4-12}
         & & & Best of non-LLM & GPT-3 & GPT-3.5 & GPT-4 & GPT-4o & Table-GPT & \jellyfish-7B & \jellyfish-8B & \jellyfish-13B \\ \hline
         \multirow{4}{*}{ED} & \multirow{2}{*}{Seen} & Adult & \underline{99.10} & \underline{99.10} & 92.01 & 92.01 & 83.58 & -- & 77.40  & 73.74 & \textbf{99.33} \\
         & & Hospital & 94.40 & \textbf{97.80} & 90.74 & 90.74 & 44.76 & -- & 94.51  & 93.40 & \underline{95.59} \\ \cline{2-12}
         & \multirow{2}{*}{Unseen} & Flights & 81.00 & -- & -- & \textbf{83.48} & 66.01 & -- & 69.15 & 66.21 & \underline{82.52} \\
         & & Rayyan & 79.00 & -- & -- & \underline{81.95} & 68.53 & -- & 75.07 & 81.06 & \textbf{90.65} \\ \hline
         \multirow{4}{*}{DI} & \multirow{2}{*}{Seen} & Buy & 96.50 & 98.50 & 98.46 & \textbf{100} & \textbf{100} & -- & 98.46 & 98.46 & \textbf{100} \\
         & & Restaurant & 77.20 & 88.40 & \underline{94.19} & \textbf{97.67} & 90.70 & -- & 89.53 & 87.21 & 89.53 \\ \cline{2-12}
         & \multirow{2}{*}{Unseen} & Flipkart & 68.00 & -- & -- & \textbf{89.94} & 83.20 & -- & 87.14 & \underline{87.48} & 81.68 \\
         & & Phone & 86.70 & -- & -- & \textbf{90.79} & 86.78 & -- & 86.52 & 85.68 &\underline{87.21} \\ \hline
         \multirow{3}{*}{SM} & \multirow{2}{*}{Seen} & MIMIC-III & 20.00 & -- & -- & 40.00 & 29.41 & -- & \textbf{53.33} & \underline{45.45} & 40.00 \\
         & & Synthea & 38.50 & 45.20 & \underline{57.14} & \textbf{66.67} & 6.56 & -- & 55.56 & 47.06 & 56.00 \\ \cline{2-12}
         & Unseen & CMS & \underline{50.00} & -- & -- & 19.35 & 22.22 & -- & 42.86 & 38.10 & \textbf{59.29} \\ \hline
         \multirow{8}{*}{EM} & \multirow{6}{*}{Seen} & Amazon-Google & 75.58 & 63.50 & 66.50 & 74.21 & 70.91 & 70.10 & \textbf{81.69} & \underline{81.42} & 81.34 \\
         & & Beer & 94.37 & \textbf{100} & 96.30 & \textbf{100} & 90.32 & 96.30 & \textbf{100.00} & \textbf{100.00} & 96.77 \\
         & & DBLP-ACM & \textbf{98.99} & 96.60 & 96.99 & 97.44 & 95.87 & 93.80 & 98.65 & 98.77 & \underline{98.98} \\
         & & DBLP-GoogleScholar & \underline{95.70} & 83.80 & 76.12 & 91.87 & 90.45 & 92.40 & 94.88 & 95.03 & \textbf{98.51} \\
         & & Fodors-Zagats & \textbf{100} & \textbf{100} & \textbf{100} & \textbf{100} & 93.62 & \textbf{100} & \textbf{100} & \textbf{100} & \textbf{100} \\
         & & iTunes-Amazon & 97.06 & \underline{98.20} & 96.40 & \textbf{100} & 98.18 & 94.30 & 96.30 & 96.30  & 98.11 \\ \cline{2-12}
         & \multirow{2}{*}{Unseen} & Abt-Buy & 89.33 & -- & -- & \textbf{92.77} & 78.73 & -- & 86.06 & 88.84 & \underline{89.58} \\
         & & Walmart-Amazon & 86.89 & 87.00 & 86.17 & \textbf{90.27} & 79.19 & 82.40 & 84.91 & 85.24 & \underline{89.42} \\ \hline
         \multicolumn{3}{|c|}{Average} & 80.44 & - & - & \underline{84.17} & 72.58 & - & 82.74 & 81.55 & \textbf{86.02} \\
         \hline
    \end{tabular}
    }
    \label{tab:exp:seen-tasks}
\end{table*}

\subsection{DP Performance}
\myparagraph{Seen Tasks}
Table~\ref{tab:exp:seen-tasks} reports the performance on the seen tasks. GPT-4 performs the best in most cases (11 out of 19). However, its score on the CMS dataset of SM is mediocre. \jellyfish-13B wins the second most (7 out of 19) and reports the best average score due to advantage over GPT-4 on the CMS dataset. Comparing \jellyfish-13B with GPT-3, GPT-3.5, GPT-4o, and Table-GPT, \jellyfish-13B wins in more cases. \jellyfish-13B also outperforms best of non-LLMs on all unseen datasets and all but one seen datasets. Note that for non-LLM methods, all the datasets are seen because they need to be fine-tuned on them. Meanwhile, the 7B and 8B \jellyfish models also exhibit competitiveness, especially for DI and EM, and their average scores surpass best of non-LLMs and GPT-4o. 

\begin{table}[!t]
    \small
    \centering
    \caption{Precision (P), recall (R), and F1 score on SM.}
    \begin{tabular}{|c|c|c|c|c|c|c|c|c|c|c|}
        \hline
        \multirow{2}{*}{\textbf{Type}} & \multirow{2}{*}{\textbf{Dataset}} & \multicolumn{9}{c|}{\textbf{Model}} \\ \cline{3-11}
        & & \multicolumn{3}{c|}{SMAT} & \multicolumn{3}{c|}{GPT-4} & \multicolumn{3}{c|}{\jellyfish-13B} \\ \hline
        & & P & R & F1 & P & R & F1 & P & R & F1 \\ \hline
        \multirow{2}{*}{Seen} & MIMIC-III & 11.5 & \textbf{84.6} & 20.2 & 33.33 & 50.0 & \textbf{40.0} & \textbf{45.45} & 35.71 & \textbf{40.0} \\
        & Synthea & 24.4 & \textbf{90.9} & 38.5 & \textbf{71.42} & 62.5 & \textbf{66.67} & 41.18 & 87.50 & 56.00 \\ \hline
        Unseen & CMS & 33.9 & \textbf{95.0} & 50.0 & \textbf{60.0} & 11.5 & 19.35 & 57.14 & 61.54 & \textbf{59.26} \\
        \hline
    \end{tabular}
    \label{tab:exp:sm-detail}
\end{table}

\begin{table*}[!t]
    \small
    \centering
    \caption{DP performance on unseen tasks, micro-F1 for CTA and F1 for AVE. CTA is a seen task for RoBERTa. Zero-shot is used for LLMs. ``--'' indicates numbers not reported in prior works for this dataset.}
    \resizebox{\linewidth}{!}{   
    \begin{tabular}{| c | c | p{.1\linewidth}<{\centering} | p{.1\linewidth}<{\centering} | p{.12\linewidth}<{\centering} | p{.08\linewidth}<{\centering} | p{.08\linewidth}<{\centering} | p{.08\linewidth}<{\centering} | p{.08\linewidth}<{\centering} | p{.08\linewidth}<{\centering} | p{.08\linewidth}<{\centering} | p{.08\linewidth}<{\centering} |}
        \hline
        \multirow{2}{*}{\textbf{Task}} & \multirow{2}{*}{\textbf{Dataset}} & \multicolumn{10}{c|}{\textbf{Model}} \\ \cline{3-12}
        & & RoBERTa (159 shots) & RoBERTa (356 shots) & Stable Beluga 2 70B & SOLAR 70B & GPT-3.5 & GPT-4 & GPT-4o & \jellyfish-7B & \jellyfish-8B & \jellyfish-13B \\ \hline
        CTA & SOTAB & 79.20 & 89.73 & -- & -- & \underline{89.47} & \textbf{91.55} & 65.06 & 83.00 & 76.33 & 82.00 \\ \hline
        \multirow{2}{*}{AVE} & AE-110k & -- & -- & 52.10 & 49.20 & \textbf{61.30} & 55.50 & 55.77 & 56.09 & \underline{59.55} & 58.12 \\ 
        & OA-Mine & -- & -- & 50.80 & 55.20 & \underline{62.70} & \textbf{68.90} & 60.20 & 51.98 & 59.22 & 55.96 \\
        \hline
    \end{tabular}
    }
    \label{tab:exp:unseen-tasks}
\end{table*}

Among the four tasks, SM is the hardest task, and all the competitors report relatively low F1 score. Looking into the datasets, we find that even humans have difficulties in telling whether the two  attributes match, given only name and description. To compare the methods in more detail, we report precision and recall in Table~\ref{tab:exp:sm-detail}. The non-LLM method, SMAT, reports the highest recall, yet with a very low precision. Among its results, only 1 out of 3 -- 9 is true positive. This iss because many SM-tailored methods seek high recall, in order to find more candidates for further verification. \jellyfish-13B exhibits relatively high precision (41\% -- 57\%), and is close to GPT-4 on the unseen dataset of CMS. This suggests that \jellyfish-13B can be used as a verification method (1 out of 2 is true positive) on top of a filtering approach (e.g., SMAT). 
\myparagraph{Unseen Tasks}
Table~\ref{tab:exp:unseen-tasks} reports the performance comparison on the unseen tasks. For CTA, GPT-4 performs the best. \jellyfish models also exhibit competitiveness, especially for the 7B and 13B models. For AVE, all the \jellyfish models showcase strong generalizability. In particular, \jellyfish-8B and \jellyfish-13B surpass the two 70B models on both datasets, and outperform GPT-4 on the AE-110k dataset. 

\begin{table}[!t]
    \small
    \centering
    \caption{Impact of instruction tuning for DP on the unseen task of CTA. ``+ task'' denotes the model tuned for the task.}
    \begin{tabular}{|c|c|c|c|c|c|}
        \hline
        OOP2-13B & + ED & + DI & + SM & + EM & \jellyfish-13B \\ \hline
        56.40 & 74.20 & 79.20 & 76.70 & 71.50 & \textbf{82.00} \\ 
        \hline
    \end{tabular}
    \label{tab:exp:impact-seen-cta}
\end{table}

\begin{table}[!t]
    \small
    \centering
    \caption{Impact of prompt engineering on the unseen task of CTA, varying options in stages and chain-of-thought (CoT) over \jellyfish-13B.}
    \begin{tabular}{| p{.2\linewidth}<{\centering} | p{.2\linewidth}<{\centering} | p{.2\linewidth}<{\centering} | p{.2\linewidth}<{\centering} |}
        \hline
        One-stage, w/o CoT & One-stage, w/ CoT & Two-stage, w/o CoT & Two-stage, w/ CoT \\ \hline
        51.50 & 58.00 & 67.00 & \textbf{82.00} \\ 
        \hline
    \end{tabular}
    \label{tab:exp:impact-prompt-cta}
\end{table}

To drill down to the impact of tuning on unseen tasks, we investigate the case of CTA with \jellyfish-13B as an example. Table~\ref{tab:exp:impact-seen-cta} helps us find out which task contributes the most to this unseen task. When tuning with only one task, the model reports a micro-F1 in the range of 71\% -- 79\%, with DI being the highest. We suppose this is because DI is exactly the inverse operation of CTA, i.e., DI fills the value of an attribute, and meanwhile CTA infers the type of an attribute given a set of sample values. Moreover, the four tasks jointly contributes to an overall micro-F1 of 82\% and it surpasses the performance of tuning with only DI, showcasing the usefulness of other tasks as well.

Further, we conduct an ablation study to study the impact of prompting and report the results in Table~\ref{tab:exp:impact-prompt-cta}. The two-stage pipeline performs better than the one-stage pipeline, and chain-of-thought, which splits the inference of column types into four steps, is also useful, in line with the observation in a previous work~\cite{korini2023column}. This demonstrates that the prompt engineering techniques developed for existing LLM-based solutions also work for \jellyfish-13B. In doing so, the design of prompts for \jellyfish-13B on unseen tasks is rendered much easier, as users may directly follow those used in existing works. 

\subsection{Improvement of \jellyfish over Base Models}
\label{sec:exp:base-models}
Table~\ref{tab:exp:base-models} compares \jellyfish models and their base models on DP tasks. Consistent performance improvement is observed on all datasets for the 7B and 13B models, and on all but one dataset for the 8B model. The improvement of the 7B model is the most significant, with an average score of 35. For the 8B and 13B models, the improvement is also remarkable, with an average of 18 and 21, respectively. We also observe that the tuning benefits the performance on unseen datasets and unseen tasks. Such impact is the most significant on EM's unseen data, showcasing that the knowledge injected through tuning applies well to unseen scenarios. 

\begin{table*}[!t]
    \small
    \centering
    \caption{Improvement of \jellyfish over base models on DP. Zero-shot is used on seen datasets and few-shot is used on unseen datasets. All datasets are unseen for base models.}
    \resizebox{\linewidth}{!}{
    \begin{tabular}{| c | c | c | p{.1\linewidth}<{\centering} | p{.14\linewidth}<{\centering} | p{.1\linewidth}<{\centering} | p{.14\linewidth}<{\centering} | p{.1\linewidth}<{\centering} | p{.14\linewidth}<{\centering} |}
         \hline
         \multirow{2}{*}{\textbf{Task}} & \multirow{2}{*}{\textbf{Type}} & \multirow{2}{*}{\textbf{Dataset}} & \multicolumn{6}{c|}{\textbf{Model}} \\ \cline{4-9}
         & & & Mistral-7B & \jellyfish-7B & Llama 3-8B & \jellyfish-8B & OOP2-13B & \jellyfish-13B \\ \hline
         \multirow{4}{*}{ED} & \multirow{2}{*}{Seen} & Adult & 20.66 & 77.40 (+56.74) & 47.42 & 73.74 (+26.32) & 61.53 & 99.33 (+37.80) \\
         & & Hospital & 37.09 & 94.51 (+57.42) & 52.51 & 93.40 (+40.89) & 63.24 & 95.59 (+32.35) \\ \cline{2-9}
         & \multirow{2}{*}{Unseen} & Flights & 28.07 & 69.15 (+41.08) & 67.71 & 66.21 (-1.50) & 73.01 & 82.52 (+9.51) \\
         & & Rayyan & 22.86 & 75.07 (+52.21) & 62.46 & 81.06 (+18.64) & 89.37 & 90.65 (+1.28) \\ \hline
         \multirow{4}{*}{DI} & \multirow{2}{*}{Seen} & Buy & 76.92 & 98.46 (+21.54) & 86.15 & 98.46 (+12.31) & 89.23 & 100 (+10.77) \\
         & & Restaurant & 18.75 & 89.53 (+70.78) & 43.02 & 87.21 (+44.19) & 81.40 & 89.53 (+8.13) \\ \cline{2-9}
         & \multirow{2}{*}{Unseen} & Flipkart & 79.52 & 87.14 (+7.62) & 66.50 & 87.48 (+20.98) & 78.49 & 81.68 (+3.19)\\
         & & Phone & 76.72 & 86.52 (+9.80) & 82.16 & 85.68 (+3.52) & 84.33 & 87.21 (+2.88) \\ \hline
         \multirow{3}{*}{SM} & \multirow{2}{*}{Seen} & MIMIC-III & 6.90 & 53.33 (+46.43) & 14.81 & 45.45 (+30.64) & 36.36 & 40 (+3.64)\\
         & & Synthea & 26.67 & 55.56 (+28.89) & 23.52 & 47.06 (+23.54) & 22.22 & 56 (+33.78) \\ \cline{2-9}
         & Unseen & CMS & 0 & 42.86 (+42.86) & 27.02 & 38.10 (+11.08) & 13.33 & 59.29 (+45.96) \\ \hline
         \multirow{8}{*}{EM} & \multirow{6}{*}{Seen} & Amazon-Google & 36.51 & 81.69 (+45.15) & 60.67 & 81.42 (+20.75) & 36.70 & 81.34 (+44.64) \\
         & & Beer & 69.57 & 100 (+30.43) & 88 & 100 (+12) & 85.71 & 96.77 (+11.06) \\
         & & DBLP-ACM & 85.30 & 98.65 (+13.35) & 82.14 & 98.77 (+16.63) & 78.86 & 98.98 (+20.12) \\
         & & DBLP-GoogleScholar & 59.54 & 94.88 (+35.34) & 76.15 & 95.03 (+18.88) & 59.48 & 98.51 (+39.03) \\
         & & Fodors-Zagats & 66.67 & 100 (+33.33) & 95.23 & 100 (+4.77) & 92.68 & 100 (+7.32) \\
         & & iTunes-Amazon & 70.97 & 96.30 (+25.33) & 79.36 & 96.30 (+16.94) & 57.45 & 98.11 (+40.66) \\ \cline{2-9}
         & \multirow{2}{*}{Unseen} & Abt-Buy & 36.99 & 86.06 (+49.07) & 44.60 & 88.84 (+44.24) & 31.51 & 89.58 (+58.07)\\
         & & Walmart-Amazon & 63.14 & 84.91 (+21.77) & 59.69 & 85.24 (+25.55) & 65.21 & 89.42 (+24.21)\\ \hline
         \multirow{1}{*}{CTA} & \multirow{1}{*}{Unseen} & SOTAB & 23.49 & 83.00 (+59.1) & 64.25 & 76.33 (+12.08) & 56.36 & 82.00 (+25.64) \\ \hline
         \multirow{2}{*}{AVE} & \multirow{2}{*}{Unseen} & AE-110k & 32.92 & 56.09 (+23.17) & 56.33 & 59.55 (+3.22) & 43.87 & 58.12 (+14.25) \\
         & & OA-Mine & 32.44 & 51.98 (+19.54)& 55.57 & 59.22 (+3.65) & 54.81 & 55.96 (+1.15) \\ \hline
         \multicolumn{3}{|c|}{Average} & 44.17 & 80.14 (+35.97) & 60.69 & 79.30 (+18.60) & 61.60 & 83.21 (+21.61) \\
         \hline
    \end{tabular}
    }
    \label{tab:exp:base-models}
\end{table*}

\subsection{Impact of Knowledge Injection}
To evaluate the impact of knowledge injection, we report in Table~\ref{tab:exp:knowledge-injection} the results for OOP2-13B and its tuned version with knowledge either injected or not. Comparing OOP2-13B and the one without injected knowledge, the performance is significantly raised on seen datasets but drops on a few unseen datasets. When we turn on knowledge injection, the performance further improves the performance on seen datasets and the improvement is consistent on all but two datasets. Furthermore, the improvement is also observed and more significant on unseen datasets, because like seen datasets of Amazon-Google and Beer, they are also product data. This observation suggests that the domain knowledge learned through tuning indeed enhances the model's generalizability to unseen datasets. In addition, the impact is the most remarkable on CMS, the unseen dataset of SM, remedying the model's performance on this dataset and making it highly competitive. 

\begin{table}[!t]
    \small
    \centering
    \caption{Impact of knowledge injection, zero-shot. ``w/o KNL'' denotes the model tuned without injected knowledge in the DP task data.}
    \begin{tabular}{| c | c | c | p{.14\linewidth}<{\centering} | p{.2\linewidth}<{\centering} | p{.14\linewidth}<{\centering} |}
         \hline
         \multirow{2}{*}{\textbf{Task}} & \multirow{2}{*}{\textbf{Type}} & \multirow{2}{*}{\textbf{Dataset}} & \multicolumn{3}{c|}{\textbf{Model}} \\ \cline{4-6}
         & & & OOP2-13B & \jellyfish-13B (w/o KNL) & \jellyfish-13B \\ \hline
         \multirow{4}{*}{ED} & \multirow{2}{*}{Seen} & Adult & 61.53 & 72 & \textbf{99.33} \\
         & & Hospital & 63.24 & 69.81 & \textbf{95.59} \\ \cline{2-6}
         & \multirow{2}{*}{Unseen} & Flights & 73.01 & 65.44 & \textbf{82.52} \\
         & & Rayyan & 89.37 & 76.14 & \textbf{90.65} \\ \hline
         \multirow{4}{*}{DI} & \multirow{2}{*}{Seen} & Buy & 89.23 & 93.85 & \textbf{100} \\
         & & Restaurant & 81.40 & 88.37 & \textbf{89.53} \\ \cline{2-6}
         & \multirow{2}{*}{Unseen} & Flipkart & 78.49 & \textbf{82.80} & 81.68 \\
         & & Phone & 84.33 & 83.58 & \textbf{87.21} \\ \hline
         \multirow{3}{*}{SM} & \multirow{2}{*}{Seen} & MIMIC-III & 36.36 & \textbf{46.15} & 40 \\
         & & Synthea & 22.22 & 53.33 & \textbf{56} \\ \cline{2-6}
         & Unseen & CMS & 13.33 & 7.14 & \textbf{59.29} \\ \hline
         \multirow{8}{*}{EM} & \multirow{6}{*}{Seen} & Amazon-Google & 36.70 &  77.78 & \textbf{81.34} \\
         & & Beer & 85.71 &  93.33 & \textbf{96.77} \\
         & & DBLP-ACM & 78.86 &  97.36 & \textbf{98.98} \\
         & & DBLP-GoogleScholar & 59.48 &  93.10 & \textbf{98.51} \\
         & & Fodors-Zagats & 92.68 & \textbf{100} & \textbf{100} \\
         & & iTunes-Amazon & 57.45 & 93.10 & \textbf{94.55} \\ \cline{2-6}
         & \multirow{2}{*}{Unseen} & Abt-Buy & 31.51 & 86.29 & \textbf{89.58} \\
         & & Walmart-Amazon & 65.21 & 74.15 & \textbf{89.42} \\
         \hline
    \end{tabular}
    \label{tab:exp:knowledge-injection}
\end{table}

\begin{table*}[!t]
    \small
    \centering
    \caption{NLP performance on the Open LLM Leaderboard.}
    \resizebox{\linewidth}{!}{
    \begin{tabular}{|c|c|p{.12\linewidth}|p{.14\linewidth}|p{.12\linewidth}|p{.14\linewidth}|p{.12\linewidth}|p{.12\linewidth}|p{.12\linewidth}|}
        \hline
        \multirow{2}{*}{\textbf{Size}} & \multirow{2}{*}{\textbf{Model}} & MMLU & WinoGrande & ARC & TruthfulQA & GSM8K & HellaSwag & Average \\ 
        & & (5-shot) & (0-shot) & (25-shot) & (0-shot) & (8-shot) & (10-shot) & \\
        \hline
        \multirow{2}{*}{7B} & Mistral-7B & 62.91 & 73.88 & 63.48 & 66.91 & 41.32 & 84.79 & 65.55 \\ 
        & \jellyfish-7B & 62.08 (-0.83) & 72.69 (-1.19) & 63.48 (+0.00) & 64.76 (-2.15) & 37.91 (-3.41) & 84.48 (-0.31) & 64.23 (-1.32) \\ \hline
        \multirow{2}{*}{8B} &Llama 3-8B & 64.51 & 71.74 & 61.01 & 51.63 & 70.36 & 78.61 & 66.31 \\
        & \jellyfish-8B & 64.23 (-0.28) & 72.06 (+0.32) & 60.15 (-0.14) & 51.83 (+0.20) & 69.29 (-1.07) & 77.92 (-0.69) & 65.76 (-0.56) \\ \hline
        \multirow{2}{*}{13B} & OOP2-13B & 54.49 & 74.03 & 62.63 & 52.56 & 25.32 & 83.24 & 58.71 \\ 
        & \jellyfish-13B & 53.04 (-1.45) & 74.19 (+0.16) & 62.88 (+0.25) & 52.56 (+0.00) & 24.26 (-1.06) & 83.16 (-0.08) & 58.35 (-0.36) \\         
        \hline
    \end{tabular}
    }
    \label{tab:exp:nlp}
\end{table*}

\subsection{NLP Performance}
Table~\ref{tab:exp:nlp} compares \jellyfish models and their original models on various NLP benchmarks~\cite{hendrycks2020measuring,sakaguchi2021winogrande,lin2021truthfulqa,chollet2019measure,cobbe2021training,zellers2019hellaswag} of the Open LLM Leaderboard~\cite{open-llm-leaderboard}. For the 8B and 13B models, their NLP performance roughly retains after tuning for DP, with very slight change (0.56 and 0.36 on average, respectively), and even improves on two benchmarks. This is because we use natural language instructions to tune Jellyfish for DP tasks, keeping the same interaction format of their base models. The 7B model sacrifices more NLP performance (1.32 on average) for DP performance. We think this reflects the no free lunch theorem~\cite{wolpert1997no}, considering its smallest size among the three. 

\subsection{Impact of Instruction Data Configuration}
\label{sec:exp:impact-configuration}
We study the impact of the data configuration in the instruction data. For this set of experiments, we randomly sample data from the datasets in Table~\ref{tab:datasets-dptuning} and disable the data preparation techniques regarding positives and missing values (Section~\ref{sec:tuning:prepare}) to see the impact of dataset size clearly. 

\begin{figure}[!t]
  \centering
  \includegraphics[width=1\linewidth]{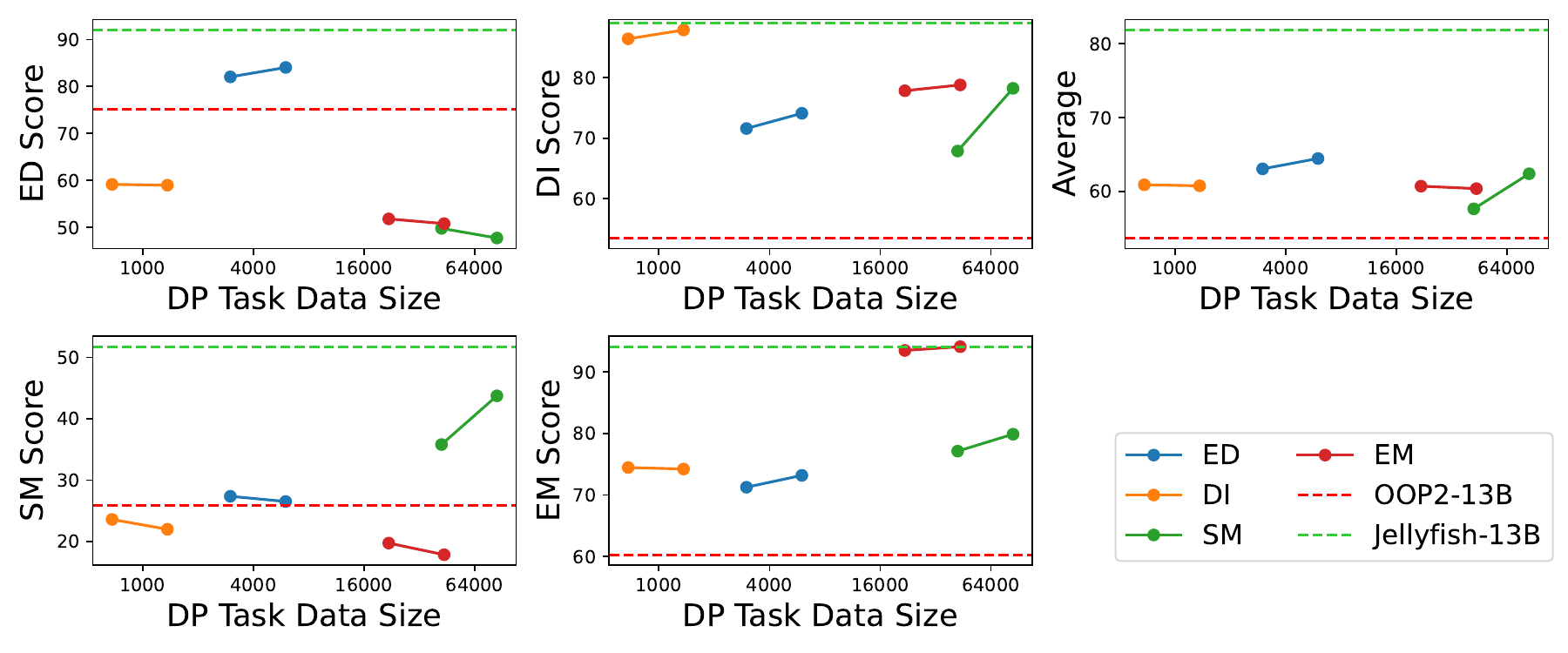}
  \caption{Impact of tuning with single-task data on DP performance, zero-shot. Above red line is positive.}
  \label{fig:impact-single-dp-dp}
\end{figure}

\begin{figure*}[!t]
  \centering
  \includegraphics[width=\linewidth]{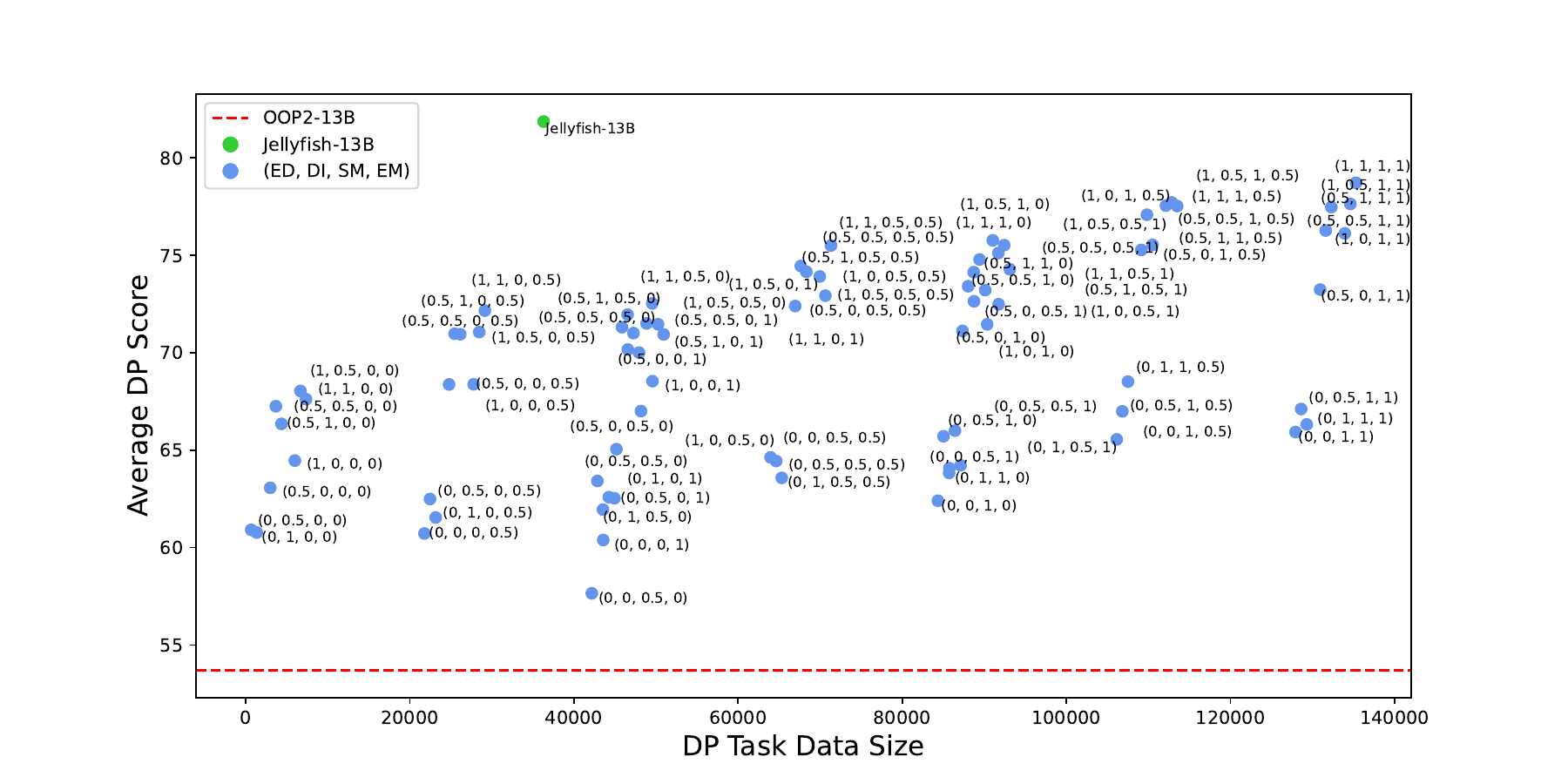}
  \caption{Impact of tuning with multi-task data on DP performance. Numbers in parenthesis indicate the percentage of data used for each task.}
  \label{fig:impact-multi-dp-dp}
\end{figure*}

To simplify the evaluation, we tune the 13B model with data for a single DP task and evaluate its effect. By varying the amount of data, Figure~\ref{fig:impact-single-dp-dp} displays how the tuning data for a specific task affects the DP performance. In general, the four tasks are all useful in improving the overall performance. For intra-task performance (e.g., ED to ED), as expected, the tuning data has a significantly positive impact. For inter-task performance, ED and SM are generally positive to other tasks, while DI and EM report negative effects. Such impact on the overall DP performance is also observed when we increase the amount of tuning data (e.g., doubling EM from 21k to 43k). We also find that DI can benefit from all the other three tasks. We think this is because the other three tasks all contain correct values for the attributes, thereby enhancing the model's ability in filling missing values. In addition, the benefit of increasing tuning data for SM is obvious. Overall, these observations results in the data configuration in constructing \jellyfish (Section~\ref{sec:tuning:prepare}). 

We study the impact of tuning the 13B model with multi-task data and plot the results in Figure~\ref{fig:impact-multi-dp-dp}. By feeding the tuning set with data for more tasks, it is obvious that they jointly contribute to better DP performance, and the improvement is consistent. When fully utilized the data, as indicated by (1, 1, 1, 1), it achieves the best performance. Based on the above results, we construct the \jellyfish data by appropriately choosing the size of data for each task. Moreover, with the data preparation techniques (Section~\ref{sec:tuning:prepare}) applied, \jellyfish-13B, even with less amount of tuning data, performs better than (1, 1, 1, 1) in Figure~\ref{fig:impact-multi-dp-dp}. 

\begin{figure*}[!t]
  \centering
  \includegraphics[width=\linewidth]{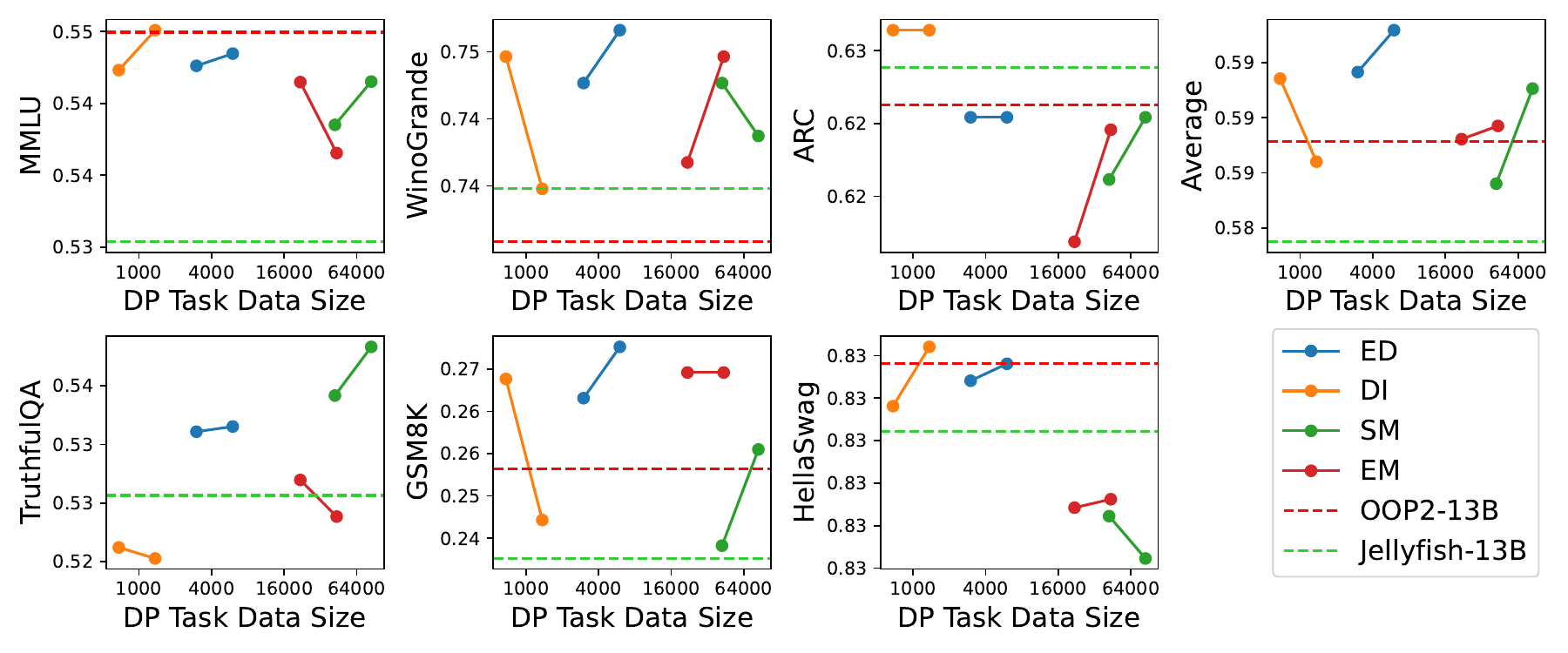}
  \caption{Impact of tuning with single-task data on NLP performance. Above red line is positive.}
  \label{fig:impact-single-dp-nlp}
\end{figure*}

\begin{figure*}[!t]
  \centering
  \includegraphics[width=\linewidth]{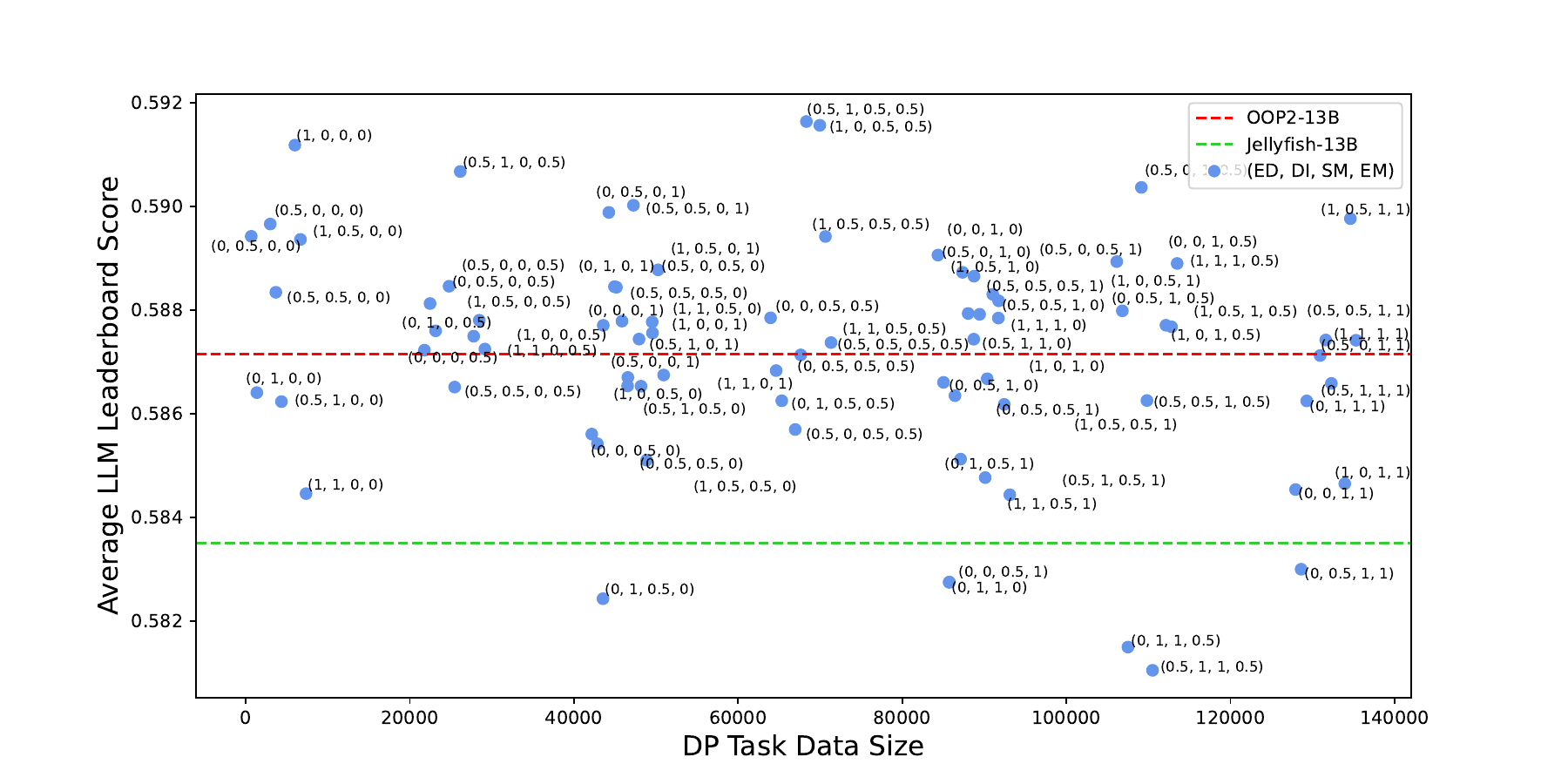}
  \caption{Impact of tuning with multi-task data on NLP performance. Numbers in parenthesis indicate the percentage of data used for each task.}
  \label{fig:impact-multi-dp-nlp}
\end{figure*}

Then, we evaluate how the data for a specific DP task affects the NLP performance and report the results in Figure~\ref{fig:impact-single-dp-nlp}. In general, ED and EM exhibit positive impacts on the overall NLP performance. By increasing the amount of tuning data, all the tasks, except DI, are positive to NLP tasks. Specifically, SM turns from negative to positive when the dataset size is doubled, whereas the trend for DI is reversed, resulting in a significant drop. To drill down to each benchmark, all the four tasks are positive to WinoGrande, while they are generally negative to MMLU, and neutral to the other benchmarks, roughly in line with the results in Table~\ref{tab:exp:nlp}. This experiment indicates that we need to choose an appropriate data size for each DP task, specifically, with moderately less data for DI, to prevent the model from losing its NLP capability. 

We also test the impact of tuning the 13B model with multi-task data on its NLP performance over the six benchmarks used in Table~\ref{tab:exp:nlp}. The results are reported in Figure~\ref{fig:impact-multi-dp-nlp}. The general trend is that with data for more tasks, the NLP performance has a drop, yet this change, as shown in more sporadic points, is less consistent than what we observed in Figure~\ref{fig:impact-multi-dp-dp}. It is noteworthy that the overall decrease in NLP performance is moderate, with an average of 0.36 (from 58.71 to 58.35) for \jellyfish-13B. 

\subsection{Impact of Reasoning Data}
\label{sec:exp:impact-reasoning}
\begin{figure}[!t]
  \centering
  \includegraphics[width=1\linewidth] {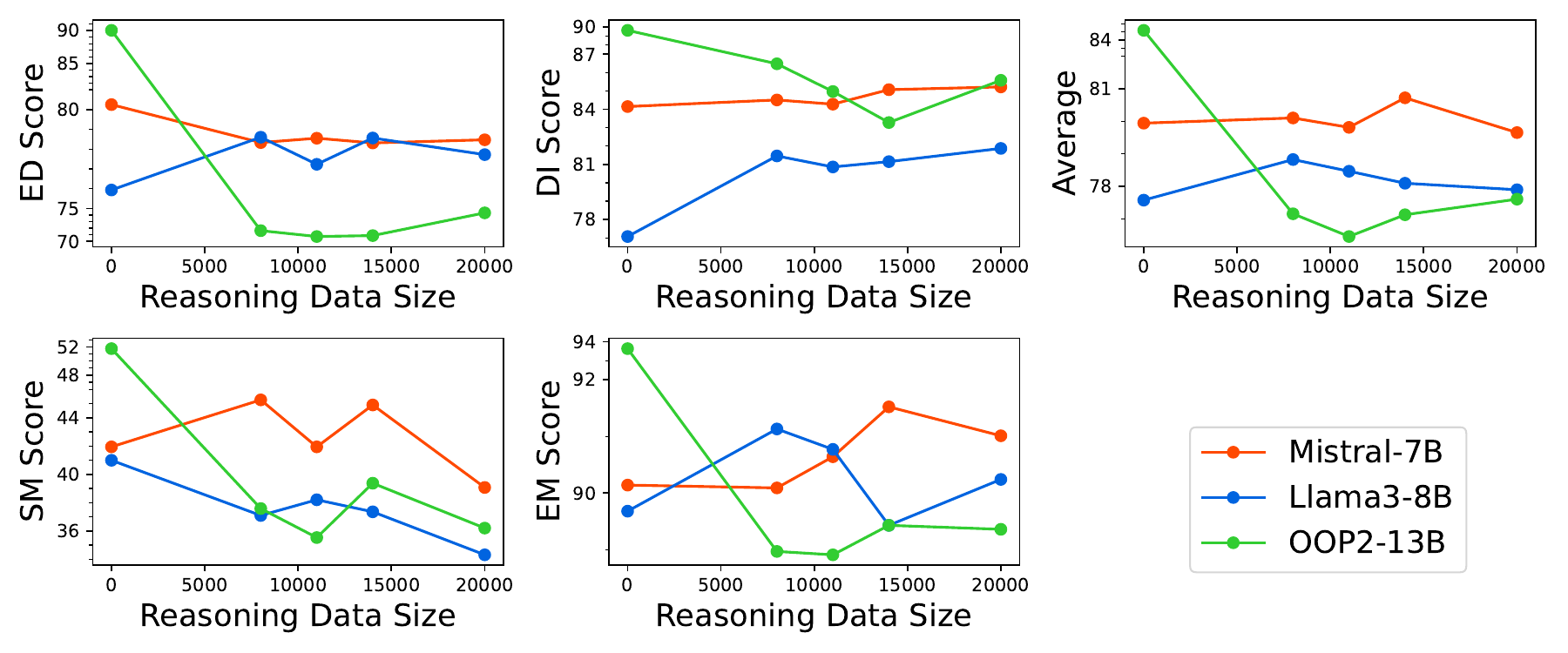}
  \caption{Impact of reasoning data on DP performance, zero-shot, plotted in log scale to show trends clearly.}
  \label{fig:reasoning-data}
\end{figure}

Figure~\ref{fig:reasoning-data} shows how reasoning data, varying from 8k, 11k, 14k, to 20k instances, impacts the DP performance. For the 7B and 8B models, the average scores increase first and then drop when more reasoning data is used for tuning, suggesting that small amount of reasoning data -- with the rationale behind DP -- can enhance the model's DP performance. Seeing this, we choose 14k and 8k for the two models, respectively, as the reasoning data size for tuning, striking a balance for the overall performance. For the 13B model, the scores drastically reduce and then rebound with more reasoning data. This may be attributed to the reasoning and logic capabilities of OOP2-13B, which are intended to enhance those of Llama 2 but ultimately do not align well with the underlying logic of DP. Only when the DP reasoning data reaches 20k, the model learns to handle DP well with reasoning. Nonetheless, the scores are still below those without reasoning, and thus we choose not to tune the 13B model with reasoning data. 

\subsection{Evaluation of Interpretation}
\label{sec:exp:interpretation}
We evaluate the performance of \jellyfish's 7B and 8B models and compare them with GPT-3.5 (\texttt{gpt-3.5-turbo-0613}). Given an answer output by \jellyfish, we generate reasons using both \jellyfish and GPT-3.5, and request GPT-4o to decide which one is better. Note that GPT-4o is unaware of the correct answer to the question in DP. As such, it needs to judge by its own analysis of the question as well. 

\begin{table}[!t]
    \small
    \centering
    \caption{Head-to-head comparison of GPT-3.5 and \jellyfish-7B/8B on interpretation, judged by GPT-4o. The two comparisons share the same sets of questions and the same answers from GPT-3.5.}
    \begin{tabular}{| c | c | p{.14\linewidth}<{\centering} | p{.14\linewidth}<{\centering} | p{.14\linewidth}<{\centering} | p{.14\linewidth}<{\centering} |}
         \hline
         \multirow{2}{*}{\textbf{Task}} & \multirow{2}{*}{\textbf{Dataset}} & \multicolumn{2}{c|}{\textbf{Comparison 1}} & \multicolumn{2}{c|}{\textbf{Comparison 2}} \\ \cline{3-6}
         & & GPT-3.5 & \jellyfish-7B & GPT-3.5 & \jellyfish-8B\\ \hline
         \multirow{2}{*}{ED} & Adult & \textbf{17} & 3 & 4 & \textbf{16}\\
         & Hospital & 4 & \textbf{16} & 4 & \textbf{16}\\ \hline
         \multirow{2}{*}{DI} & Buy & 4 & \textbf{16} & 4 & \textbf{16}\\
         & Restaurant & \textbf{10} & \textbf{10} & 9 & \textbf{11}\\  \hline
         \multirow{1}{*}{SM} & Synthea & \textbf{15} & 5 & 3 & \textbf{17} \\ \hline
         \multirow{6}{*}{EM} & Amazon-Google & 3 & \textbf{17} & 2 & \textbf{18} \\
         & Beer & \textbf{13} & 7 & 7 & \textbf{13}\\
         & DBLP-ACM & \textbf{11} & 9 & 2 & \textbf{18} \\
         & DBLP-GoogleScho8lar & \textbf{16} & 4 & 9 & \textbf{11} \\
         & Fodors-Zagats & \textbf{13} & 7 & \textbf{13} & 7 \\
         & iTunes-Amazon & \textbf{12} & 8 & 2 & \textbf{18} \\ \hline
         \multicolumn{2}{|c|}{Total} & \textbf{118} & 102 & 59 & \textbf{161} \\ \hline
         \multicolumn{2}{|c|}{Winning Rate} & \textbf{53.63\%} & 46.36\% & 26.81\% & \textbf{73.18\%} \\ \hline
    \end{tabular}
    \label{tab:exp:interpretation}
\end{table}

Table~\ref{tab:exp:interpretation} reports the head-to-head comparison of the two models. The results show that \jellyfish-7B is almost on a par with GPT-3.5, while \jellyfish-8B is a much better interpreter. In particular, \jellyfish-7B wins over GPT-3.5 on 3 out of 11 datasets, reporting an overall winning rate of 46\% across 220 tests. \jellyfish-8B defeats GPT-3.5 on 10 out of 11 datasets, achieving a winning rate of 73\%. 
To delve into detailed comparison of \jellyfish-8B and GPT-3.5, we show an example on the Amazon-Google dataset in EM. The prompt is as follows.

\begin{prompt}{Head-to-Head Instance}
    You are an AI assistant that follows instruction extremely well. The user will give you a question. Your task is to answer as faithfully as you can. While answering, provide detailed explanation and justify your answer.\\
    Answer your reason first, then finish your response in a separate line with and ONLY with your final answer.\\
    Product A: [name: "adobe dreamweaver cs3 [ mac ]", manufacturer: "adobe", price: "399.0"]\\
    Product B: [name: "adobe cs3 web premium", price: "1659.99"]\\
    Are Product A and Product B the same Product?
\end{prompt}

The reason given by \jellyfish-8B is as follows.
\begin{snippet}
    No, Product A and Product B are not the same product. Although both products are manufactured by Adobe, they have different names and prices. Product A is named "adobe dreamweaver cs3 [mac]" and is priced at \$399.00. On the other hand, Product B is named "adobe cs3 web premium" and is priced at \$1659.99. The difference in price and name suggests that these are two distinct products with different features and functionalities.\\
    In addition, the name of Product A includes the specific software name "dreamweaver cs3" and the operating system it is designed for, "mac". This suggests that Product A is a specific version of the Dreamweaver software for Mac users. In contrast, Product B is named "cs3 web premium", which suggests that it is a more comprehensive package that includes multiple Adobe software products, rather than just a single software application.\\
    Therefore, based on the available information, it is clear that Product A and Product B are not the same product.\\
    Final Answer: No, Product A and Product B are not the same product.",
\end{snippet}

The reason given by GPT-3.5 is as follows.
\begin{snippet}
    No, Product A and Product B are not the same product.
    The reason is that although both products are from Adobe and are part of the CS3 suite, they have different names and prices. Product A is specifically named "Adobe Dreamweaver CS3 [Mac]" and is priced at \$399.0, while Product B is named "Adobe CS3 Web Premium" and is priced at \$1659.99. The difference in names and prices indicates that they are different products within the CS3 suite, likely with different features and functionalities.\\
    Therefore, Product A and Product B are not the same product.\\
    Final answer: No.
\end{snippet}

GPT-4o compares the outputs of both models in the following aspects, where Model 1 refers to \jellyfish-8B and Model 2 refers to GPT-3.5: 
\begin{inparaenum} [(1)]
    \item Faithfulness to Instruction: Model 1: Provides a detailed explanation, comparing the names, prices, and inferred functionalities of both products. It also elaborates on the specific nature of Product A (Dreamweaver for Mac) versus the more comprehensive package suggested by Product B (Web Premium). Model 2: Also states that the products are different, citing differences in names and prices. However, it provides less detail and does not delve into the possible functionalities or the specific nature of the products as much as Model 1 does.
    \item Justification: Model 1: Justifies the conclusion by explaining the implications of the product names and prices. It infers that Product A is a specific software for Mac users, while Product B is a more comprehensive suite, which is a logical and well-supported argument. Model 2: Justifies the conclusion by pointing out the differences in names and prices but does not provide as thorough an analysis of what those differences imply about the products.
    \item Clarity and Completeness: Model 1: The response is clear, detailed, and logically structured. It ensures that the user understands why the products are different. Model 2: The response is clear but lacks the depth and completeness found in Model 1's explanation.
    \item Conclusion: Model 1: Concludes with a clear and well-supported final answer. Model 2: Concludes with a clear final answer but lacks the depth of reasoning provided by Model 1.
    \item Winner: Model 1. 
\end{inparaenum}
Based on the analysis, GPT-4o decides that \jellyfish-8B performed better because it adhered more strictly to the original instruction by providing a clear, concise response without unnecessary repetition. In addition, with its built-in knowledge, \jellyfish-8B pointed out the difference in functionalities, whereas GPT-3.5 merely described the difference on the surface.  

Furthermore, we find out that when reasoning, GPT-3.5 even fails to respond with a correct answer of matching or not for EM, as shown in its mediocre performance on datasets like Amazon-Google and DBLP-GoogleScholar in Table~\ref{tab:exp:seen-tasks}. In contrast to the above example of \jellyfish-8B's landslide win, GPT-3.5 only has a slight edge when it wins. For instance, in an example of the Amazon-Google dataset, GPT-4o points out that GPT-3.5 has more focused justification and additional insights into the implications of the differences between the products, yet it also mentioned that GPT-3.5's repetition of the final answer is a minor deviation from the instruction's format.

\subsection{Evaluation of Efficiency}
\label{sec:exp:efficiency}
With 8 GPUs of A100 80G, instruction tuning spends around 5 hours for \jellyfish-13B, 3 hours for \jellyfish-7B and \jellyfish-8B. For inference on single GPU of A100 80G, \jellyfish-7B, 8B, and 13B spend 0.07, 0.08, and 0.15 seconds, respectively, on average to process an instance. As a reference, GPT-4 spends an average of 1 -- 8 seconds per instance. Although LLMs require substantial computational resources, thereby increasing the cost of use and compromising the efficiency, some non-LLM methods, such as RoBERTa and those built upon it (e.g., IPM), need fine-tuning when applied to unseen datasets. This fine-tuning time should be counted towards total time expense for fair comparison. Moreover, advanced learning techniques enables \jellyfish models to be quantized~\cite{liu2023llm} or distilled to improve efficiency, which will be considered in the future. To further save processing time, users are also suggested using a simple but faster method to retrieve a set of candidates and then apply Jellyfish models to the candidates. For example, for EM, blocking is often used to group similar records together based on certain attributes and narrow the comparisons to within each block.

For batch processing of multiple instances, the speed can be improved by 1.31 and 1.27 times for 8B and 13B models, respectively, when prefix caching is enabled in vLLM. However, this optimization is not available for the 7B model due to the sliding window attention used in Mistral-7B.

As for memory consumption, Jellyfish-7B, 8B, and 13B spend 18GB, 20GB, and 30GB VRAM (including the model), respectively. To further reduce memory consumption, we can resort to activation-aware weight quantization~\cite{lin2024awq}. By doing so, the memory consumption of the 7B and 8B models can be reduced to 7.5GB and 8GB, respectively, without compromising much of the performance (-1.25 and -0.52 average micro-F1/accuracy for the 7B and 8B models, respectively).
\section{Related Works}
\label{sec:related}
Since works on LLMs have been introduced in Section~\ref{sec:prelim:llm}, we briefly review related works on DP here.

\myparagraph{Seen Tasks}
The tasks targeted in this paper collectively form the most critical part of DP, and they have been extensively studied.
\begin{itemize} 
  \item \textbf{ED:} Traditional methods mainly depend on hand-crafted rules~\cite{chu2013holistic}, pattern discovery~\cite{chu2015katara}, outlier detection~\cite{prokoshyna2015combining}, or statistical modeling~\cite{huang2018auto,wang2019uni}. Recent works employ more advanced ML techniques such as few-shot learning based on a noisy channel model (HoloDetect)~\cite{heidari2019holodetect}, or resort to a series of ML pipelines (Raha)~\cite{mahdavi2019raha}, including feature engineering, clustering, and classification. 
  \item \textbf{DI:} While rule-based solutions~\cite{rekatsinas2017holoclean,song2018enriching} remain one of the prevalent approaches, another stream of works develops ML models for this task, including variational autoencoders~\cite{nazabal2020handling}, generative adversarial networks~\cite{yoon2018gain}, and attention mechanisms~\cite{wu2020attention,tihon2021daema}. To seek better imputation performance, recent progress utilizes PLMs to capture semantics~\cite{mei2021capturing}.
  \item \textbf{SM:} The use of similarity matrices is a traditional way~\cite{sagi2013schema}. More advanced methods utilize ML techniques~\cite{gal2019learning}, including deep learning models~\cite{shraga2020adnev}. SMAT~\cite{zhang2021smat} is an approach leveraging attention-based deep learning. A recent attempt employs GPT-4 for SM~\cite{sheetrit2024rematch}. 
  \item \textbf{EM:} The procedure is divided into blocking and in-block pairwise matching for the sake of efficiency. Blocking groups pairs of entities that potentially match into the same block, and then pairwise matching is performed within each block to find matching entities. Traditional solutions for blocking mostly rely on attribute equivalence, hashes, or similarities~\cite{papadakis2020blocking}. Recently, the feasibility of using DL methods for blocking has also been examined~\cite{thirumuruganathan2021deep}, following the use of DL for pairwise matching~\cite{mudgal2018deep}. In addition, there are tools that handle both steps such as Megallan~\cite{konda2016magellan} and Ditto~\cite{li2020deep}. A recent evaluation validates the effectiveness of in-context learning in enhancing LLMs' EM performance~\cite{peeters2023entity}. 
\end{itemize}

\myparagraph{Unseen Tasks}
We review the related studies on CTA and AVE. 
\begin{itemize} 
  \item \textbf{CTA:} As a typical table understanding task, it often appears in the studies on table representation learning~\cite{iida2021tabbie,deng2022turl,suhara2022annotating}. These approaches fine-tune PLMs, typically BERT~\cite{devlin2018bert} and its variants~\cite{jiao2019tinybert,liu2019roberta}. Recently, ChatGPT has been utilized to solve this task~\cite{korini2023column}. 
  \item \textbf{AVE:} Early approaches employ LSTM-CRF~\cite{kozareva2016recognizing,zheng2018opentag}. With the prevalence of PLMs, like CTA, many solutions to AVE resort to using BERT~\cite{xu2019scaling,wang2020learning,zhu2020multimodal}. A recent work~\cite{brinkmann2023product} considered fine-tuning GPT-3.5 and prompting GPT-4, and compared with open-source LLMs like Stable Beluga 2~\cite{stable-beluga-2} and SOLAR~\cite{solar}. 
\end{itemize}

\myparagraph{Generic Solution} Whereas the above solutions are specialized for a task, recent progress developed generic solutions to DP based on GPT-3~\cite{narayan2022can}, or GPT-3.5 and GPT-4~\cite{zhang2023large}, basically employing various prompt engineering techniques on frozen LLMs. Fine-tuning GPT-3.5 and ChatGPT for a variety of table-related tasks has also been investigated~\cite{li2023table}, and several DP tasks are covered. 
  
\myparagraph{Other DP Tasks}
Besides the ones covered by this paper, there are many other DP tasks. We name a few examples. 
\begin{itemize} 
  \item Data repairing corrects erroneous values in a dataset. Typical solutions are HoloClean~\cite{rekatsinas2017holoclean} and Baran~\cite{mahdavi2020baran}. HoloClean can detect errors and perform repairing subsequently. Baran only repair errors and resort to Raha to detect errors. Recent advancements~\cite{lew2021pclean,qin2023bclean} utilized Bayesian inference to capture dependencies between attributes. 
  \item Data fusion is the process of integrating multiple data sources that contain information about the same set of entities, with possibly conflicting attribute values. Surveys of early attempts are available~\cite{li2015truth,canalle2021survey}, with a detailed comparison of various fusion methods on deep web data~\cite{li2015truth}. More recent endeavors targeted multi-truth data fusion~\cite{azzalini2023enhancing} and golden record~\cite{heidari2023record}.
  \item Data transformation is the process of converting data from one format into another format. Notable approaches are transformation by user-specified examples~\cite{he2018transform} and learning from large collections of paired table columns~\cite{jin2020auto}. In addition, the aforementioned generic DP solution also covers this task~\cite{narayan2022can}.
\end{itemize}

\myparagraph{Data Preparation}
DP is also studied in the name of data preparation, which manipulates raw data into a form that can be readily analyzed. A notable Python library is DataPrep~\cite{dataprep}. In addition to the DP tasks listed above, data augmentation~\cite{chepurko2020arda,miao2021rotom,liu2021adaptive,zhao2022leva} is another key operation in data preparation. Another line of work studies dataset discovery~\cite{bogatu2020dataset,koutras2021valentine,fan2022semantics,nargesian2022data}, particularly for integrating data lake tables~\cite{khatiwada2022integrating} where joinable~\cite{dong2022deepjoin},  unionable~\cite{khatiwada2023santos}, and related table search~\cite{zhang2020finding} are often used for identifying candidates. Despite search speed being a key concern, LLMs are anticipated to be used on top of their outcomes for automated integration in a data lake~\cite{arora2023language}.
\section{Conclusions}
\label{sec:concl}
\myparagraph{Contributions}
We studied the problem of instruction-tuning LLMs as universal DP task solvers. By devising data preparation and knowledge injection techniques, we proposed \jellyfish, which enables users to craft instructions for DP tasks. Another notable feature of \jellyfish is its interpretation ability, providing explanations of its outputs. We tuned three base models ranging from 7B to 13B, which can operate on a local GPU without compromising data security. The experiments demonstrated the competitiveness of \jellyfish against existing DP solutions, impressive generalizability to new tasks, the ability of retaining performance in NLP tasks, as well as the competence in interpretation.  

\myparagraph{Limitations}
We investigated six DP tasks, whereas there are still many other tasks (e.g., data repairing, data fusion, and data transformation). 

We discovered that our reasoning data compromises the 13B model's DP performance, possibly because  OpenOrca-Platypus2-13B's reasoning and logic do not align well with the underlying logic of DP. In contrast, the 7B and 8B models, derived from native models Mistral-7B and Llama 3-8B, respectively, can benefit from the use of reasoning data. Due to the lack of (approximately) 13B size for these two base models, we cannot deliver better DP performance than the 13B model while preserving the interpretation ability. Nonetheless, we believe that our instruction data can apply to more advanced base models, with which better DP performance and interpretation ability could be both achieved. 

Another limitation is that our prompt is designed as instance-based, rather than the table-based setting which was adopted in many non-LLM approaches. This is partially due to the token limitation of the LLMs we used (e.g., 4096 tokens for a 13B model), and compromises efficiency when we use our models for large-scale datasets. In our future work, we will consider designing prompts that take multiple instance or a table at a time. 

Furthermore, when using our models for practical data mining pipelines, we also need to carefully consider the issues of preprocessing and postprocessing. For example, the input data may be scanned copies and contain hierarchical tables, while we focus on relational tables in this work.  

\myparagraph{Future Work} 
Our future work aims to address the limitations outlined above, in particular, expanding the instruction data of \jellyfish to encompass more DP tasks, such as data repairing and data transformation. Furthermore, our future research directions include the development of a quantized or distilled model to enhance processing speed, as well as a multi-agent system for adaptable, conversational, code-free DP pipeline. 

\section*{Acknowledgements}
This work is mainly supported by NEC Corporation and partially supported by JSPS Kakenhi 23K17456, 23K25157, 23K28096, and JST CREST JPMJCR22M2. 


\bibliographystyle{abbrv}
\bibliography{references-dp,references-nlp}

\clearpage

\onecolumn

\appendix

\section{Experimental Setup}
\label{sec:app:exp-setup-detail}

\myparagraph{Hyperparameters}
The following hyperparameters are used for \jellyfish models' tuning and inference:

\begin{table}[h!]
  \small
  \centering
  \caption{Hyperparameter setting.}
  \begin{tabular}{|l|l|l|}
    \hline
    \textbf{Category} & \textbf{Parameter} & \textbf{Value} \\ \hline
    \multirow{7}{*}{Tuning} & lora\_target & q\_proj, k\_proj, v\_proj, o\_proj \\ \cline{2-3} 
    & per\_device\_train\_batch\_size & 2 \\ \cline{2-3} 
    & gradient\_accumulation\_steps & 2 \\ \cline{2-3} 
    & learning\_rate & 3e-5 \\ \cline{2-3} 
    & num\_train\_epochs & 5 \\ \cline{2-3} 
    & lora\_rank & 32 \\ \cline{2-3} 
    & lora\_alpha & 32 \\ \hline
    \multirow{3}{*}{Inference} & temperature & 0.35 \\ \cline{2-3} 
    & top\_p & 0.9 \\ \cline{2-3} 
    & top\_k & 20 \\ \hline
  \end{tabular}
\end{table}

\section{Instruction Data Prompts}
\label{sec:app:prompts}
\subsection{DP Task Data}
For DP task data, we show the prompt for each task, using one dataset as an example. Then, we show the prompt for reasoning data, which slightly differs from DP task data. The prompts for inference are the same as tuning, except that dataset-specific knowledge is optional. The prompts for reasoning ground truth collection and head-to-head judge are used for Mixtral. 

We use \jellyfish-13B's prompts as examples. For other models, users may adjust them according to the prompt format of their base models (e.g., using ``[INST] [/INST]'' blocks for the 7B model).

\begin{prompt}{DP Task Data -- ED (Adult)}
  \textbf{(system message)} You are an AI assistant that follows instruction extremely well. User will give you a question. Your task is to answer as faithfully as you can.\\
  \textbf{(task description)} Your task is to determine if there is an error in the value of a specific attribute within the whole record provided. The attributes may include age, workclass, education, marital status, occupation, relationship, race, sex, hours per week, country, and income.\\
  \textbf{(injected knowledge)} Errors may include, but are not limited to, spelling errors, inconsistencies, or values that don't make sense given the context of the whole record.\\
  \textbf{(instance content)} Record [age: "18-21", workclass: "Private", education: "Some-college", maritalstatus: "Never-married", occupation: "Other-service", relationship: "Own-child", race: "White", sex: "Male", hoursperweek: "30", country: "United-States", income: "eLssThan50K"]\\
  Attribute for Verification: [income: "eLssThan50K"]\\
  \textbf{(question)} Is there an error in the value of the "income" attribute?\\
  \textbf{(output format)} Choose your answer from: [Yes, No]
\end{prompt}

\begin{prompt}{DP Task Data -- DI (Restaurant)}
  \textbf{(system message)} You are an AI assistant that follows instruction extremely well. User will give you a question. Your task is to answer as faithfully as you can.\\
  \textbf{(task description)} You are presented with a restaurant record that is missing a specific attribute: the city. Your task is to deduce or infer the city of the restaurant using the available information in the record. You may be provided with fields like 'Name', 'Address', 'Phone', and 'Type' to help you in the inference.\\
  \textbf{(instance content)} Record: [name: "darbar", addr: "44 w. 56th st.", phone: "212-432-7227", type: "indian"].\\
  \textbf{(question)} Based on the provided restaurant record, what would you infer is the value for the missing attribute "City"?\\
  \textbf{(output format)} Answer the name of the city.
\end{prompt}

\begin{prompt}{DP Task Data -- SM (MIMIC-III)}
  \textbf{(system message)} You are an AI assistant that follows instruction extremely well. User will give you a question. Your task is to answer as faithfully as you can.\\
  \textbf{(task description)} Your task is to determine if the two attributes (columns) are semantically equivalent in the context of merging two tables. Each attribute will be described by its name and a brief description. Your goal is to assess if they refer to the same information based on these names and descriptions provided.\\
  \textbf{(instance content)} Attribute A is [name: "visit\_occurrence-visit\_end\_date", description: "the end date of the visit. if this is a one-day visit the end date should match the start date."].\\
  Attribute B is [name: "admissions-dischtime", description: "dischtime provides the date and time the patient was discharged from the hospital."].\\
  \textbf{(question)} Are Attribute A and Attribute B semantically equivalent?\\
  \textbf{(output format)} Choose your answer from: [Yes, No]
\end{prompt}

\begin{prompt}{DP Task Data -- EM (Beer)}
  \textbf{(system message)} You are an AI assistant that follows instruction extremely well. User will give you a question. Your task is to answer as faithfully as you can.\\
  \textbf{(task description)} You are tasked with determining whether two Products listed below are the same based on the information provided. Carefully compare all the attributes before making your decision.\\
  \textbf{(injected knowledge)} Note that missing values (N/A or "nan") should not be used as a basis for your decision.\\
  \textbf{(instance content)} Product A: [name: "Sequoia American Amber Ale", factory: "Wig And Pen"]\\
  Product B: [name: "Aarhus Cains Triple A American Amber Ale", factory: "Aarhus Bryghus"]\\
  \textbf{(question)} Are Product A and Product B the same Product?\\
  \textbf{(output format)} Choose your answer from: [Yes, No]
\end{prompt}

\subsection{Reasoning Data}
\label{sec:app:prompts:reasoning}
To construct reasoning data, we use the following prompt. 
\begin{prompt}{Reasoning Data}
  \textbf{(system message)} \textbf{[same as DP Task Data]} While answering, provide detailed explanation and justify your answer.\\
  \textbf{(task description -- question)} \textbf{[same as DP Task Data]}\\
  \textbf{(output format)} After your reasoning, finish your response in a separate line with and ONLY with your final answer. Choose your final answer from [Yes, No].
\end{prompt}

We use the following prompt to collect ground truth from Mixtral (for the Beer dataset in EM).
\begin{prompt}{Reasoning Ground Truth Collection}
  \textbf{(system message -- output format)} \textbf{[same as Reasoning Data]}\\
  \textbf{(injected knowledge)} Note that different factories can belong to the same parent company. The company name of Product B may occur in its product name.\\
  \textbf{(answer hint)} You can use the "Hint" below, but your response cannot contain any information from it.\\
  Hint: the final answer is "No"
\end{prompt}

In the above prompt, we inject a piece of knowledge specific to the dataset. With this additional knowledge, Mixtral can produce high-quality reasoning result. Note that such knowledge is not prompted to \jellyfish models, as for unseen datasets such knowledge is not always available. In addition to the injected knowledge, Mixtral also receives a hint to the answer of yes or no. As such, we can guarantee that the output reason always points to the correct direction. A sample answer from Mixtral is given below.

\begin{snippet}
  To determine if Product A and Product B are the same, we need to compare their names and factories.
  The names of the products are "Sequoia American Amber Ale" and "Aarhus Cains Triple A American Amber Ale". While both are American Amber Ales, the names are not identical, with Product B having additional words "Aarhus Cains Triple A". Therefore, based on the name, we cannot conclude that the products are the same.
  Next, let's consider the factories. Product A is produced at "Wig And Pen" while Product B is produced at "Aarhus Bryghus". These are different factories, and the fact that they may belong to the same parent company or use similar production processes does not change the fact that the products are made at different locations.
  Therefore, based on the comparison of the names and factories, we can conclude that Product A and Product B are not the same product.\\
  Final answer: No
\end{snippet}

\section{Injected Knowledge}
\label{sec:app:injected-knowledge}

\begin{table}[H]
    \small
    \centering
    \caption{General knowledge.}    
    \begin{tabular}{|p{.95\linewidth}|}
        \hline
        \textbf{Prompt} \\ \hline
        Missing values (N/A or "nan") should not be used as a basis for your decision. \\ \hline
        If there are missing values, you should make inferences based only on the information that is available.\\
        \hline
    \end{tabular}
    \label{tab:knowledge:general}
\end{table}

\begin{table}[H]
    \small
    \centering
    \caption{Task-specific knowledge.}    
    \begin{tabular}{|c|p{.9\linewidth}|}
         \hline
         \textbf{Task} & \textbf{Prompt} \\ \hline
         \multirow{5}{*}{ED} & Errors may include, but are not limited to, spelling errors, inconsistencies, or values that don't make sense given the context of the whole record. (Used when showing the whole record) \\ \cline{2-2}
         & Errors can include, but are not limited to, spelling errors, inconsistencies, or values that don't make sense for that attribute. (Used when showing only one attribute) \\ \cline{2-2}
         & Capitalization should not be a factor in deciding whether there is an error or not. \\ \hline
         \multirow{4}{*}{DI} & Note that values such as 'nan' and 'N/A' mean missing vaules, and they are not considered as errors. \textbf{(used when we decide not to treat missing values as errors)} \\ \cline{2-2}
         & Note that values such as 'nan' and 'N/A' mean missing values, and they ARE errors. \textbf{(used when we decide to treat missing values as errors)} \\ \hline
         EM & To determine if two values are identical, you need to examine both their full names and corresponding acronyms. \\
         \hline
    \end{tabular}
    \label{tab:knowledge:task}
\end{table}

\begin{table}[H]
    \small
    \centering
    \caption{Dataset-specific knowledge.}    
    \begin{tabular}{|c|c|p{.6\linewidth}|}
         \hline
         \textbf{Task} & \textbf{Dataset} & \textbf{Prompt} \\ \hline
         \multirow{5}{*}{ED} & \multirow{4}{*}{Adult} & Both the 'age' attribute and the 'hoursperweek' attribute can represent a range of integer values. \\ \cline{3-3}
         & & Verify the consistency of target attribute with related attributes to identify any errors. \\ \cline{2-3}
         & Hospital & The value of attribute "score" can be a percentage number. \\ \hline
         \multirow{2}{*}{DI} & \multirow{2}{*}{Restaurant} & The city can often be deduced from the area code of the phone number and the specific street name. \\ \hline
         \multirow{14}{*}{EM} & \multirow{3}{*}{Amazon-Google} & Different editions, versions, or operating systems for the same software are all considered as different products. \\ \cline{3-3}
         & & You should compare the two product numbers first. \\ \cline{2-3}
         & \multirow{4}{*}{Beer} & Note that different factories can belong to the same parent company. \\ \cline{3-3}
         & & Beverages that undergo different production processes, such as the use of various types of wood in the barrelling process, may be considered distinct products. \\ \cline{2-3}
         & Fodors-Zagats & The type of a specific restaurant might vary between different datasets. \\ \cline{2-3}
         & \multirow{2}{*}{iTunes-Amazon} & The length of the same song might vary slightly across different datasets due to rounding or data entry discrepancies. \\ \cline{2-3}
         & \multirow{2}{*}{DBLP-ACM} & The names of authors might be presented in various formats or sequences, even when referring to the same article. \\ \cline{2-3}
         & \multirow{2}{*}{DBLP-GoogleScholar} & The names of authors might be presented in various formats or sequences, even when referring to the same article. \\ 
         \hline
    \end{tabular}
    \label{tab:knowledge:dataset}
\end{table}

\section{Few-Shot Prompting}
\label{sec:app:few-shot}
We apply few-shot prompting by manually selecting a subset of data instances from the dataset and labeling them. For instance, a few-shot example for the Beer dataset is presented as follows:

\begin{prompt}{Few-Shot Prompting}
  \textbf{(system message -- injected knowledge)} \textbf{[same as DP Task Data]}\\
  \textbf{(1st example's instance content)} \#\#\# Instruction: Product A: [name: "Shirt Tail Amber", factory: "Iron Hill Brewery \& Restaurant"]\\
  Product B: [name: "Iron Hill Shirt Tail Amber", factory: "Iron Hill Maple Shade"]\\
  \textbf{(1st example's question)} Are Product A and Product B the same Product?\\
  \textbf{(1st example's output format)} Choose your answer from: [Yes, No]\\
  \textbf{(1st example's answer)} \#\#\# Response: Yes\\
  \textbf{(other examples)} ...\\
  \textbf{(instance content -- output format)} \textbf{[same as DP Task Data]} \#\#\# Response:
\end{prompt}

The example follows the same format of instance content, question, and output format as in the DP task data. It also provides the answer indicated by \textit{\#\#\# Response: Yes}. Whereas we only show an positive example here, it is suggested to include both positive and negative examples. After the final example, the instance to be processed is presented in the prompt, and the model follows the same output format as demonstrated in the examples. 

Since ground truths are usually not available in real applications, users can handcraft few-shot examples for inference. On the other hand, few-shot examples can be automatically generated by randomly injecting errors for ED and DI, such as missing values, typographical/formatting errors, and randomly swapping values for two columns in a tuple or two tuples in a column. For SM and EM, we can employ rule-based methods (e.g., blocking rules~\cite{konda2016magellan}) to quickly find a few matches and use them as few-shot examples. 

\subsection{Error Detection}
The few-shot examples for the Flights and Rayyan datasets are given as follows.

\begin{prompt}{Flights  -- 1st Example}
  \#\#\# Instruction:\\
  Record [datasource: "flightview", flight: "AA-3063-SLC-LAX", scheduled departure time: "nan", actual departure time: "8:40 p.m.", scheduled arrival time: "nan", actual arrival time: "9:11 p.m."]\\
  Attribute for Verification: [scheduled departure time: "nan"]\\
  Question: Is there an error in the value of the "scheduled departure time" attribute?\\
  Choose your answer from: [Yes, No]\\
  \#\#\# Response:\\
  Yes
\end{prompt}

\begin{prompt}{Flights  -- 2st Example}
  \#\#\# Instruction:\\
  Record [datasource: "aa", flight: "AA-3823-LAX-DEN", scheduled departure time: "9:00 p.m.", actual departure time: "nan", scheduled arrival time: "12/02/2011 12:15 a.m.", actual arrival time: "nan"]\\
  Attribute for Verification: [scheduled arrival time: "12/02/2011 12:15 a.m."]\\
  Question: Is there an error in the value of the "scheduled arrival time" attribute?\\
  Choose your answer from: [Yes, No]\\
  \#\#\# Response:\\
  Yes
\end{prompt}

\begin{prompt}{Flights  -- 3rd Example}
  \#\#\# Instruction:\\
  Record [datasource: "flightview", flight: "AA-616-DFW-DTW", scheduled departure time: "9:05 a.m.", actual departure time: "10:11 a.m.", scheduled arrival time: "12:35 p.m.", actual arrival time: "1:18 p.m."]\\
  Attribute for Verification: [datasource: "flightview"]\\
  Question: Is there an error in the value of the "datasource" attribute?\\
  Choose your answer from: [Yes, No]\\
  \#\#\# Response:\\
  No
\end{prompt}

\begin{prompt}{Rayyan  -- 1st Example}
  \#\#\# Instruction:\\
  Record [article\_title: "A re-appraisal of screening for colour vision impairments", article\_language: "nan", journal\_title: "Child: Care, Health \& Development", jounral\_abbreviation: "nan", journal\_issn: "0305-1862", article\_jvolumn: "23", article\_jissue: "5", article\_jcreated\_at: "1/1/97", article\_pagination: "391-398", author\_list: "{"D. M. B. Hall","E. Holroyd"}"]\\
  Attribute for Verification: [jounral\_abbreviation: "nan"]\\
  Question: Is there an error in the value of the "jounral\_abbreviation" attribute?\\
  Choose your answer from: [Yes, No]\\
  \#\#\# Response:\\
  Yes
\end{prompt}

\begin{prompt}{Rayyan  -- 2nd Example}
  \#\#\# Instruction:\\
  Record [article\_title: "Nurturing students' interest in primary care research through summer training in meta-analysis.", article\_language: "eng", journal\_title: "Academic Medicine: Journal Of The Association Of American Medical Colleges", jounral\_abbreviation: "nan", journal\_issn: "1040-2446", article\_jvolumn: "76", article\_jissue: "5", article\_jcreated\_at: "5/1/01", article\_pagination: "526", author\_list: "{"L N Meurer"}"]\\
  Attribute for Verification: [article\_jissue: "5"]\\
  Question: Is there an error in the value of the "article\_jissue" attribute?\\
  Choose your answer from: [Yes, No]\\
  \#\#\# Response:\\
  No
\end{prompt}

\begin{prompt}{Rayyan  -- 3rd Example}
  \#\#\# Instruction:\\
  Record [article\_title: "Different renal toxicity profiles in the association of cyclosporine and tacrolimus with sirolimus in rats.", article\_language: "eng", journal\_title: "Nephrology, dialysis, transplantation : official publication of the European Dialysis and Transplant Association - European Renal Association", jounral\_abbreviation: "Nephrol. Dial. Transplant.", journal\_issn: "1460-2385", article\_jvolumn: "23", article\_jissue: "10", article\_jcreated\_at: "10/1/08", article\_pagination: "3111-9", author\_list: "{"N\textbackslash{}u033cria Lloberas","Marcel la Franquesa","Josep M Cruzado","Josep M Griny\textbackslash{}ufffd\_","In\textbackslash{}u0329s Rama","Gabriela Alperovich","Immaculada Herrero-Fresneda","Joan Torras","Pepita Gim\textbackslash{}u0329nez-Bonaf\textbackslash{}u0329"}"]\\
  Attribute for Verification: [article\_pagination: "3111-9"]\\
  Question: Is there an error in the value of the "article\_pagination" attribute?\\
  Choose your answer from: [Yes, No]\\
  \#\#\# Response:\\
  Yes
\end{prompt}

\subsection{Data Imputation}
The few-shot examples for the Flikpkart and Phone datasets are given as follows.

\begin{prompt}{Flipkart  -- 1st Example}
  \#\#\# Instruction:\\
  Record: [Product Name: "Himmlisch ST381 Magnetic Sun Shade For Maruti Alto", description: "Himmlisch ST381 Magnetic Sun Shade For Maruti Alto (Side Window) Price: Rs. 1,899 Beat the heat this summer and feel like a VIP with Himmlisch Car Window Magnetic Sunshades. These magnetic sunshades create a mesh layer to stops the heat. Magnet border gets easily stick to your car window door edges (No need of Suction cups) Features: Block UV Rays Keeps Car Cool Easy to install and remove Durable and Exact Fit Provides Complete privacy Resists Heat Mesh Type Sunshade Package Contents: 1 x Set Of 4 Magnetic Sunshades,Specifications of Himmlisch ST381 Magnetic Sun Shade For Maruti Alto (Side Window) General Brand Himmlisch Model Number ST381 Magnetic Placement Position Side Window Color Black Dimensions Weight 4000 g Depth 1.1 cm In the Box Sales Package 4 Sun Shade Pack of 4"]\\
  Based on the provided product record, what would you infer is the value for the missing attribute "brand"?\\
  Answer the name of the brand.\\
  \#\#\# Response:\\
  Himmlisch
\end{prompt}

\begin{prompt}{Flipkart  -- 2nd Example}
  \#\#\# Instruction:\\
  Record: [Product Name: "dilli bazaaar Bellies, Corporate Casuals, Casuals", description: "Key Features of dilli bazaaar Bellies, Corporate Casuals, Casuals Material: Fabric Occasion: Ethnic, Casual, Party, Formal Color: Pink Heel Height: 0,Specifications of dilli bazaaar Bellies, Corporate Casuals, Casuals General Occasion Ethnic, Casual, Party, Formal Ideal For Women Shoe Details Weight 200 g (per single Shoe) - Weight of the product may vary depending on size. Heel Height 0 inch Outer Material Fabric Color Pink"]\\
  Based on the provided product record, what would you infer is the value for the missing attribute "brand"?\\
  Answer the name of the brand.\\
  \#\#\# Response:\\
  dilli bazaaar
\end{prompt}

\begin{prompt}{Flipkart  -- 3rd Example}
  \#\#\# Instruction:\\
  Record: [Product Name: "Shining Diva Alloy Yellow Gold Bangle Set", description: "Shining Diva Alloy Yellow Gold Bangle Set (Pack of 2) Price: Rs. 499 Accentuate Your Feminine Charm Wearing This Beautiful Bangle From The House Of Shining Diva. Made From Premium Quality Material, It Will Retain Its Quality And Lustre For Years To Come. This Bangle Is Lightweight And Skin Friendly. Featuring A Stylish Design And Great Finish, It Will Definitely Give Your Overall Look An Ethereal Dimension. This Bangle Will Surely Catch Your Fancy At Once. It Is Worth Investing In And Will Definitely Get You Noticed. This Bangle Comes In A Set Of Two.Accentuate Your Feminine Charm Wearing This Beautiful Bangle From The House Of Shining Diva. Made From Premium Quality Material, It Will Retain Its Quality And Lustre For Years To Come. This Bangle Is Lightweight And Skin Friendly. Featuring A Stylish Design And Great Finish, It Will Definitely Give Your Overall Look An Ethereal Dimension. This Bangle Will Surely Catch Your Fancy At Once. It Is Worth Investing In And Will Definitely Get You Noticed. This angle Comes In A Set Of Two."]\\
  Based on the provided product record, what would you infer is the value for the missing attribute "brand"?\\
  Answer the name of the brand.\\
  \#\#\# Response:\\
  Shining Diva
\end{prompt}

\begin{prompt}{Phone  -- 1st Example}
  \#\#\# Instruction:\\
  Record: Record: [Product Name: "UNLOCKED RIM BlackBerry Pearl Flip 8220 Smart Cell Phone - Red"]\\
  Based on the provided cellphone record, what would you infer is the value for the missing attribute "brand"?\\
  Answer the name of the brand.\\
  \#\#\# Response:\\
  BlackBerry
\end{prompt}

\begin{prompt}{Phone  -- 2nd Example}
  \#\#\# Instruction:\\
  Record: [Product Name: "OtterBox Apple iPhone 4 \& 4S Protective ION Defender Series Case (Retail Packaging) Black"]\\
  Based on the provided cellphone record, what would you infer is the value for the missing attribute "brand"?\\
  Answer the name of the brand.\\
  \#\#\# Response:\\
  OtterBox
\end{prompt}

\begin{prompt}{Phone  -- 3rd Example}
  \#\#\# Instruction:\\
  Record: [Product Name: "DTECH @ 2 PECES! Universal Ring Grip/Stand Holder for any Smart Device,Universal Black Bunker Ring Stand Holder for Apple iPhone 4 4s iphone 5 Samsung Galaxy s3 SIII Samsung GALAXY S6,S6 EDGE.Note II iPad 2 3 ipad mini iPod Nokia LG HTC One X etc,RING Essentials " Cell Phone and Tablets Anti Drop Ring for iPhone 6 plus iPad mini iPad2 iPad iPod Samsung GALAXY NOTE S5 Universal Mobile Devices"]\\
  Based on the provided cellphone record, what would you infer is the value for the missing attribute "brand"?\\
  Answer the name of the brand.\\
  \#\#\# Response:\\
  DTECH
\end{prompt}

\subsection{Schema Matching}
The few-shot examples for the CMS dataset are given as follows.

\begin{prompt}{CMS  -- 1st Example}
  \#\#\# Instruction:\\
  Attribute A is [name: "condition\_occurrence-condition\_source\_value", description: "the source code for the condition as it appears in the source data. this code is mapped to a standard condition concept in the standardized vocabularies and the original code is stored here for reference."]\\
  Attribute B is [name: "inpatientclaims-admtng\_icd9\_dgns\_cd", description: "claim admitting diagnosis code"]\\
  Are Attribute A and Attribute B semantically equivalent?\\
  Choose your answer from: [Yes, No]\\
  \#\#\# Response:\\
  Yes
\end{prompt}

\begin{prompt}{CMS  -- 2nd Example}
  \#\#\# Instruction:\\
  Attribute A is [name: "provider-npi", description: "the national provider identifier (npi) of the provider."]\\
  Attribute B is [name: "outpatientclaims-op\_physn\_npi", description: "operating physician – national provider identifier number"]\\
  Are Attribute A and Attribute B semantically equivalent?\\
  Choose your answer from: [Yes, No]\\
  \#\#\# Response:\\
  Yes
\end{prompt}

\begin{prompt}{CMS  -- 3rd Example}
  \#\#\# Instruction:\\
  Attribute A is [name: "visit\_detail-visit\_detail\_start\_datetime", description: "the date and time of the visit started."]\\
  Attribute B is [name: "outpatientclaims-desynpuf\_id", description: "beneficiary code"]\\
  Are Attribute A and Attribute B semantically equivalent?\\
  Choose your answer from: [Yes, No]\\
  \#\#\# Response:\\
  No
\end{prompt}

\subsection{Entity Matching}
The few-shot examples for the Abt-Buy and Walmart-Amazon datasets are given as follows.

\begin{prompt}{Abt-Buy  -- 1st Example}
  \#\#\# Instruction:\\
  Product A: [name: "samsung s3 black multimedia player yps3jab", description: "samsung s3 black multimedia player yps3jab 4 gb internal flash memory 1.8 ' tft lcd display touch-sensitive led controls multi-formats support dnse 2.0 sound engine fm tuner and recorder with presets up to 25 hours audio playback up to 4 hours video playback black finish"]\\
  Product B: [name: "samsung 4gb portable mltimdia plyr blk yps-s3jab / xaa", description: "nan"]\\
  Are Product A and Product B the same Product?\\
  Choose your answer from: [Yes, No]\\
  \#\#\# Response:\\
  Yes
\end{prompt}

\begin{prompt}{Abt-Buy  -- 2nd Example}
  \#\#\# Instruction:\\
  Product A: [name: "sony white 8 ' portable dvd player dvpfx820w", description: "sony dvp-fx820 white 8 ' portable dvd player dvpfx820w swivel \& flip screen with dual sensor for remote control control buttons on screen bezel 12 bit video dac with 108 mhz processing removable , rechargeable battery \& car adapter included white finish"]\\
  Product B: [name: "toshiba sd-p71s portable dvd player", description: "toshiba sd-p71s 7 ' portable dvd player"]\\
  Are Product A and Product B the same Product?\\
  Choose your answer from: [Yes, No]\\
  \#\#\# Response:\\
  No
\end{prompt}

\begin{prompt}{Abt-Buy  -- 3rd Example}
  \#\#\# Instruction:\\
  Product A: [name: "sony xplod 10-disc add-on cd/mp3 changer cdx565mxrf", description: "sony xplod 10-disc add-on cd/mp3 changer cdx565mxrf cd/cd-r/cd-rw and mp3 playback mp3 decoding d-bass 12-second advanced electronic shock protection fm modulator 9 modulation frequencies wireless remote"]\\
  Product B: [name: "sony cdx-565mxrf 10-disc cd/mp3 changer", description: "nan"]\\
  Are Product A and Product B the same Product?\\
  Choose your answer from: [Yes, No]
  \#\#\# Response:\\
  Yes
\end{prompt}

\begin{prompt}{Walmart-Amazon  -- 1st Example}
  \#\#\# Instruction:\\
  Product A: [name: "d-link dgs-1005g 5-port gigabit desktop switch", modelno: "dgs1005g"]\\
  Product B: [name: "d-link dgs-1005g 5-port gigabit desktop switch", modelno: "dgs-1005g"]\\
  Are Product A and Product B the same Product?\\
  Choose your answer from: [Yes, No]\\
  \#\#\# Response:\\
  Yes
\end{prompt}

\begin{prompt}{Walmart-Amazon  -- 2nd Example}
  \#\#\# Instruction:\\
  Product A: [name: "nzxt phantom crafted series atx full tower steel chassis black", modelno: "nzxt phantom"]\\
  Product B: [name: "nzxt crafted series atx full tower steel chassis - phantom white", modelno: "phantom white"]\\
  Are Product A and Product B the same Product?\\
  Choose your answer from: [Yes, No]\\
  \#\#\# Response:\\
  No
\end{prompt}

\begin{prompt}{Walmart-Amazon  -- 3rd Example}
  \#\#\# Instruction:\\
  Product A: [name: "at t prepaid gophone samsung a187 with bluetooth blue", modelno: "a187"]\\
  Product B: [name: "samsung a107 prepaid gophone at t", modelno: "a107"]\\
  Are Product A and Product B the same Product?\\
  Choose your answer from: [Yes, No]\\
  \#\#\# Response:\\
  No
\end{prompt}

\end{document}